\documentclass[lettersize,journal]{IEEEtran}

\usepackage{graphicx}
\usepackage[caption=false,font=normalsize,labelfont=sf,textfont=sf]{subfig}
\usepackage{paralist} 
\usepackage{enumitem,kantlipsum} 
\usepackage{cite}
\usepackage{bbding} 

\usepackage{algorithm} 
\usepackage[noend]{algpseudocode}

\usepackage[noend]{algpseudocode}
\usepackage{hhline}

\usepackage{color,soul} 

\usepackage{multirow} 

\usepackage{subfig} 

\usepackage[export]{adjustbox} 

\usepackage{capt-of}

\usepackage{tabularx}

\usepackage[hyphens]{url}

\usepackage{dblfloatfix}    

\usepackage{amsmath} 

\usepackage{atbegshi}
\AtBeginDocument{\AtBeginShipoutNext{\AtBeginShipoutDiscard}} 

\usepackage
{soul} 

\let\oldfootnote\footnote
\def\footnote{\ifhmode\unskip\fi\oldfootnote} 

\usepackage{nomencl} 
\makenomenclature

\begin{document}
\title{CADeSH: Collaborative Anomaly Detection\\for Smart Homes}

\author{\IEEEauthorblockN{Yair Meidan,
Dan Avraham, Hanan Libhaber, and
Asaf Shabtai}}%

\thanks{Manuscript received April 1, 2022; revised July 6, 2022; accepted July 25, 2022.}%
\thanks{This project received funding from the European Union’s Horizon 2020 research and innovation programme under grant agreement No 830927.}%
\thanks{All authors are with the Department of Software and Information Systems Engineering, Ben-Gurion University of the Negev, Beer-Sheva, Israel. Email: \{yairme, danavra, hananlib\}@post.bgu.ac.il, shabtaia@bgu.ac.il.}%
\thanks{Copyright (c) 2022 IEEE. Personal use of this material is permitted. However, permission to use this material for any other purposes must be obtained from the IEEE by sending a request to pubs-permissions@ieee.org.}

\markboth{IEEE Internet of Things Journal,~Vol.~XX, No.~XX, August~2022}%
{Meidan \MakeLowercase{\textit{et al.}}: CADeSH: Collaborative Anomaly Detection for Smart Homes}

\IEEEpubid{0000--0000/00\$00.00~\copyright~2021 IEEE}

\maketitle     

\thispagestyle{empty} 
\pagenumbering{arabic} 

\begin{abstract}
Although home IoT (Internet of Things) devices are typically plain and task oriented, the context of their daily use may affect their traffic patterns. That is, a given IoT device will probably not generate the exact same traffic data when operated by different people in different environments and when connected to different networks with different topologies and communication components. For this reason, anomaly-based intrusion detection systems tend to suffer from a high false positive rate (FPR). To overcome this, we propose a two-step collaborative anomaly detection method which first uses an autoencoder to differentiate frequent (`benign') and infrequent (possibly `malicious') traffic flows. Clustering is then used to analyze only the infrequent flows and classify them as either known ('rare yet benign') or unknown (`malicious'). Our method is collaborative, in that \emph{(1)} normal behaviors are characterized more robustly, as they take into account a variety of user interactions and network topologies, and \emph{(2)} several features are computed based on a pool of identical devices rather than just the inspected device.

We evaluated our method empirically, using 21 days of real-world traffic data that emanated from eight identical IoT devices deployed on various networks, one of which was located in our controlled lab where we implemented two popular IoT-related cyber-attacks. Our collaborative anomaly detection method achieved a macro-average area under the precision-recall curve of 0.841, an F1 score of 0.929, and an FPR of only 0.014. These promising results were obtained by using labeled traffic data from our lab as the test set, while training the models on the traffic of devices deployed outside the lab, and thus demonstrate a high level of generalizability. In addition to its high generalizability and promising performance, our proposed method also offers benefits such as privacy preservation, resource savings, and model poisoning mitigation. On top of that, as a contribution to the scientific community, our novel dataset is available online.
\end{abstract}


\section{Introduction}\label{sec:introduction}

\IEEEPARstart{I}{n} recent years, numerous cyber-attacks involving \emph{Internet of things} (IoT) devices have taken place all over the world~\cite{makhdoom2018anatomy,sengupta2020comprehensive}. In the smart home sector~\cite{arif2020investigating,lin2016iot}, these attacks include the recruitment of IoT devices to botnets~\cite{celil2020detecting} for the execution of distributed denial-of-service (DDoS) attacks~\cite{darknexus} and the utilization of IoT devices for cryptocurrency mining~\cite{bertino2017botnets,breitenbacher2019hades,log4j}. Given the consistent increase in the number of IoT devices deployed~\cite{Gartner,IDCautonomous,statista2020}, the yearly increase in the rate of IoT malware development~\cite{iotmalwaredouble} (some of which is even open-sourced~\cite{botenago}), and the stealthiness of IoT-related cyber-attacks~\cite{kaspersky2019}, it is clear why new attack detection methods are needed for risk mitigation.

\IEEEpubidadjcol

As a means of detecting IoT-related cyber-attacks, anomaly detection in network traffic data has been studied extensively~\cite{alhajri2019survey,lin2020anomaly,nomm2018unsupervised}. However, the following shortcomings are typically associated with existing methods of this kind:

\begin{itemize}[leftmargin=*]
    \item \emph{Limited generalizability.} In many cases, e.g.,~\cite{meidan2018nbaiot,pahl2018all}, model training is limited, in that just one device from each IoT model is deployed in only one network. Hence, the transferability of the trained anomaly detector to other (identical) IoT devices cannot be guaranteed. This is especially true if the monitored IoT devices are deployed in other networks and other users interact with them.
    \item \emph{Indifference to the effect of user interaction.} As noted in~\cite{cvitic2020definition}, the traffic patterns of IoT devices can be heavily influenced by a device's interaction with the user(s), other devices, and the environment. Accordingly, an IoT device's  `normal' traffic behavior captured from an arbitrary location (where, e.g., a smart streaming device user relies on just a single streaming service source and always manually changes the channel) might not generalize well to other locations (where, e.g., users rely on numerous TV content sources for their streaming device and frequently use the voice assistant to change the channel).
    \item \emph{Excessive false alarms.} Typically, anomaly detection methods only distinguish between normal and anomalous activities (i.e., a binary class label). However, when anomaly detection is used for attack detection, non-malicious anomalies could be mistakenly identified as cyber-attacks and thus lead to an excessive number of false alarms and unnecessary (automated) risk-mitigation responses. In other words, there are also legitimate reasons for traffic anomalies, such as infrequent user activities (booting a webcam or zooming in/out), or contextual changes (e.g., domestic smart motion detectors may send alerts more frequently during holidays or pandemic-related lockdowns, when families spend more time at home than usual).
\end{itemize}

To address these shortcomings, in this research we propose incorporating three enhancements for anomaly detection, especially when it is used for IoT-related attack detection in smart homes. The first enhancement is \emph{data sharing for collaborative model training}. The idea is that instead of characterizing the normal traffic behavior using just a single device from a single environment, the traffic data of multiple identical IoT devices is shared so that the training data represents a larger variety of user interactions, network topologies, bandwidths, etc. With data sharing, network- and interaction-related traffic variations can be captured, contributing to the generalization of benign traffic behaviors. Once deployed, when inspecting the first \emph{local} occurrence of a pattern which represents, e.g., a webcam zooming in, an anomaly detector  trained this way is less likely to raise a false alarm, because a similar pattern was probably already captured \emph{globally} in at least one other environment. The second proposed enhancement is \emph{collaborative feature engineering}, meaning that some features are calculated using a pool of identical devices rather than just the inspected device. For example, if an inspected device communicates with a destination IP address it never contacted before, the suspicion that such communication is malicious would decrease (leading to fewer false alarms) if \emph{other} devices have also communicated with the same destination IP address recently. Of course, this kind of feature engineering also necessitates data sharing. The third enhancement we propose is \emph{fine-tuning and extending the set of possible class labels}. Specifically, instead of dichotomously labeling each instance as either \emph{normal} or \emph{abnormal} (equivalent to \emph{benign} or \emph{malicious}, respectively, in attack detection), we consider three types of behavior and for each of them we define a corresponding attack detection-related class label, as follows:

\begin{enumerate}[leftmargin=*]
    \item \emph{Frequent (`benign').} A limited~\cite{Skowron2020} set of typical predefined operations which result in a limited number of normal traffic patterns. For instance, frequent IoT operations include sensing motion, detecting smoke, and sending video.
    \item \emph{Known (`rare yet benign').} A set of operations which occur much less frequently but are still considered normal, e.g., a webcam booting or zooming in/out. In the context of attack detection, a known event that is labeled as an anomaly is considered a false positive (FP). In our method, both \emph{frequent} and \emph{known} behaviors are eventually treated the same way, i.e, as an indication of a benign activity.
    \item \emph{Unknown (`malicious').} Extremely anomalous events which are highly indicative of the occurrence of malicious activity like a cyber-attack.
\end{enumerate}

In this work, we propose \emph{CADeSH}, a novel \emph{C}ollaborative \emph{A}nomaly \emph{De}tection method for IoT attack detection in \emph{S}mart \emph{H}omes. CADeSH is based on collaboration among IoT devices of the same model, and as such it offers several important benefits:
\begin{itemize}[leftmargin=*]
    \item \emph{Feasibility and privacy preservation.} A key benefit in using CADeSH is that instead of relying on raw network traffic data, it makes use of \emph{meta}data in IPFIX form (Internet Protocol Flow Information Export)~\cite{cisconetflowipfix,ipfixspec} which \emph{(1)} is already natively supported by most routers~\cite{verde2014no}, and \emph{(2)} is considered more privacy preserving~\cite{stalla2014porn} and less resource consuming~\cite{li2013survey} than packet-level traffic analysis approaches, such as deep packet inspection (DPI).
    \item \emph{Efficiency.} Multiple identical IoT devices are monitored simultaneously, so model retraining can be performed more frequently, based on shorter data collection periods.
    \item \emph{Quality assurance of the training data.} CADeSH includes a collaborative mechanism in which not all of the monitored traffic is used for training, but rather only instances that are widely acknowledged (i.e., by multiple identical devices) as benign and representative are used. Consequently, the likelihood of model poisoning decreases.
    \item \emph{Robustness of the trained models.} Although IoT devices are typically designed to execute a relatively small set of predefined operations~\cite{bertino2017botnets,meidan2018nbaiot}, the context of their use (in different weather conditions, by multiple people of different ages, etc.) might contribute to variability in a device's normal network traffic~\cite{cvitic2020definition}. Collaborative anomaly detection allows such traffic variations to be captured and thus contributes to the model's generalization.
\end{itemize}

In addition to the benefits listed above, we make the following contributions:

\begin{itemize}[leftmargin=*]
\item To improve the model's performance, we propose data enrichment using two additional (i.e., non-IPFIX) data sources: DNS requests and the Webroot API~\cite{webroot}.
\item For evaluation, in addition to the benign and malicious data captured in our lab, we also relied on real-world data collected from five (real) home networks over a period of 21 days. This is the entire duration of network traffic data collection. The data was provided to us by a company which provides cyber-defense services for smart homes. To enable reproducible research and facilitate future advancements in research on collaborative IoT anomaly detection, we made our (anonymized) dataset publicly available. We are not aware of other similar publicly available online datasets.%
~\footnote{https://doi.org/10.5281/zenodo.6406052}
\end{itemize}%

\begin{table*}[h]
\centering
\caption{Objectives of prior research on anomaly and attack detection, mainly in IoT settings}
\label{tab:literature_objectives}
\resizebox{\textwidth}{!}{%
\begin{tabular}{c|c|l}
\hhline{===}
\textit{\textbf{Ref.}}      & \begin{tabular}[t]{@{}c@{}}\textit{\textbf{IoT-}}\\\textit{\textbf{related}}\end{tabular} & \multicolumn{1}{c}{\textit{\textbf{Objective}}}                               \\ \hhline{===}
~\cite{meidan2018nbaiot} & \Checkmark                    & Identify DDoS attacks as soon as they are launched from IoT devices that were recruited to a botnet                     \\ \hline
~\cite{pahl2018all}         & \Checkmark                    & Create an IoT anomaly detection service                                       \\ \hline
~\cite{alrashdi2019ad}      & \Checkmark                    & Create an anomaly detection IoT system for smart cities                       \\ \hline
~\cite{asharf2020review}    & \Checkmark                    & Review machine learning (ML) and deep learning (DL) based intrusion detection systems (IDSs) used in IoT settings                                         \\ \hline
~\cite{garg2020multi}       & \Checkmark & Overcome certain limitations of the Density-Based Spatial Clustering of Applications with Noise (DBSCAN) algorithm    \\ \hline
~\cite{hasan2019attack}     & \Checkmark                    & Compare several ML algorithms for attack and anomaly detection in IoT systems \\ \hline
~\cite{markiewicz2020clust} & \Checkmark & Create an IDS for low-powered and resource-constrained IoT devices, using unsupervised learning                         \\ \hline
~\cite{mergendahl2020rapid} & \Checkmark & Create a real-time, self-training, easily deployed, anomaly-based DDoS detection system for IoT networks                \\ \hline
~\cite{nguyen2019diot}      & \Checkmark                    & Detect compromised IoT devices without sharing data                           \\ \hline
~\cite{nie2020intrusion}    &            & Improve the effectiveness of traditional ML IDSs on low-frequency attacks in high-dimensional networks                  \\ \hline
~\cite{regan2022federated} & \Checkmark                    & Propose a federated-based approach for the detection of botnet attacks, using on-device decentralized traffic data                     \\ \hline
~\cite{shahid2019anomalous} & \Checkmark                    & Create an anomaly detector for IoT network communication                     \\ \hline
~\cite{sriram2020network}   & \Checkmark                    & Create a DL-based botnet attack detector for compromised IoT devices        \\ \hline
~\cite{wani2020ddos}        & \Checkmark & Create a software-defined network (SDN) based security mechanism to detect and mitigate DDoS attacks on IoT networks  \\ \hline
~\cite{yamauchi2020anomaly} & \Checkmark                    & Detect cyber-attacks based on user behavior                                   \\ \hline
~\cite{yang2020ddos}        &                               & Create an anomaly-based DDoS attack detector, using autoencoders (AEs)                        \\ \hhline{===}
\begin{tabular}[t]{@{}c@{}}This\\paper\end{tabular} & \Checkmark & Improve the performance of IoT-related cyber-attack detection,  using a two-step collaborative anomaly detection method \\ \hhline{===}
\end{tabular}%
}
\end{table*}

\section{Research objective, scope, and assumptions}\label{sec:Research_Scope_Assumptions_and_Objectives}

\subsection{Objective and scope}\label{subsec:scope}

The main objective of this research is to develop an efficient and generalizable collaborative method for the detection of various attacks involving IoT devices, while minimizing the false positive rate (FPR).

Given the large variety of \emph{(1)} IoT application domains~\cite{perwej2019internet} and \emph{(2)} viable IoT-related attacks~\cite{sengupta2020comprehensive}, we focus on the smart home domain and propose a means of detecting two common malicious activities, namely the recruitment of devices to IoT botnets (for later execution of DDoS attacks) and unauthorized cryptocurrency mining (see Section~\ref{sec:evaluation_method}).

\subsection{Assumptions}\label{subsec:assumptions}

Considering the applicability of our proposed method, we note that collaborative attack detection (training and application) requires capturing and processing network traffic data from multiple IoT devices connected to multiple home networks. To enable this, we assume that the IoT devices are continuously monitored and their network traffic data is systematically collected into a central server. In real-world settings, this mechanism of data capturing, followed by model training and application, can be performed by Internet service providers (ISPs) as an additional countermeasure~\cite{Lastdrager2020canISPs} against IoT malware or by third-party security providers. Another assumption we make in this research is that IoT devices of the same model produce network traffic data which is relatively similar to one another because of the common intended functionality, however these traffic patterns might vary due to either \emph{(a)} differences in user interaction and network structure or \emph{(b)} cyber-attacks.

\section{Background and related work}\label{sec:background_and_related_work}

\subsection{Attack detection and anomaly detection in the IoT}\label{subsec:attack_anomaly_detection}

In recent years, traffic-based IoT attack detection methods have become more prevalent~\cite{ali2020systematic,stephens2021detecting,xing2021survey}. Many of the existing methods for IoT attack detection~\cite{diro2021comprehensive} rely on anomaly detection. In contrast to supervised methods~\cite{anthi2019supervised,su2018lightweight}, which necessitate high-quality and sufficiently-large labeled datasets for training that cover both benign traffic and known attacks, unsupervised methods typically only require benign traffic (which is relatively easy to collect), and often, they can also detect zero-day attacks~\cite{moustafa2019outlier}. The basic assumption in relying on anomaly detection for attack detection is that the accumulation of traffic data from an uninfected IoT device during normal user interaction enables normal traffic patterns of the device to be captured. Then, after modeling the normal traffic patterns using various unsupervised machine learning algorithms, extreme deviations from the norm seen in the future could indicate malicious activity.

As can be seen in Table~\ref{tab:literature_objectives}, numerous past studies utilized unsupervised machine learning algorithms for IoT attack detection. Among them, the authors of~\cite{meidan2018nbaiot} proposed training an anomaly detector for each device separately. As part of their method, outbound packets trigger the extraction of behavioral snapshots of the network, to which an anomaly detection algorithm is applied to detect DDoS attacks when they are launched from IoT devices compromised by botnets (Mirai and BASHLITE). The study compared the anomaly detection performance when using the Isolation Forest (IF), Local Outlying Factor (LOF), One-Class Support Vector Machines (OCSVM) algorithms and autoencoders (AEs) and found that the latter obtained the best detection results. Similarly, the authors of~\cite{yang2020ddos} also proposed the use of AEs to differentiate benign and malicious traffic (generated by DDoS attacks), based on a flow of packets. Unlike those two studies, in this research \emph{(a)} we train the anomaly detector using data from multiple networks (not just one), which increases generalizability, \emph{(b)} in our method, two algorithms (not just one) are applied and three class labels are considered (rather than a binary class label), and \emph{(c)} the FPR is reduced by adding the rare yet benign class label, which although considered abnormal, triggers no risk mitigating response. Unlike~\cite{meidan2018nbaiot}, our data was captured from multiple real home networks and not from just one network in a controlled lab. Like~\cite{meidan2018nbaiot}, we make our unique dataset publicly available. In the subsections that follow, we further compare our research to existing work, focusing on aspects such as the algorithmic complexity and flow, the breadth of data sources for model training, and the availability of the experimental data for reproducible research.

\begin{table*}[ht]
\centering
\caption{Algorithms in prior research on anomaly and attack detection, mainly in IoT settings}
\label{tab:literature_algorithms_collaboration}
\resizebox{0.75\textwidth}{!}{%
\begin{tabular}{c|l|c|c}
\hhline{====}
\textit{\textbf{Ref.}} &
  \multicolumn{1}{c|}{\textit{\textbf{Algorithm(s) used}}} &
  \begin{tabular}[t]{@{}c@{}}\textit{\textbf{Collaboration}}\\\textit{\textbf{approach}}\end{tabular} &
  \begin{tabular}[t]{@{}c@{}}\textit{\textbf{Shared}}\\\textit{\textbf{material}}\end{tabular} \\ \hhline{====}
~\cite{meidan2018nbaiot} & AE                                   & & \\ \hline
~\cite{pahl2018all}         & $k$-means, BIRCH (Balanced Iterative Reducing and Clustering using Hierarchies)                      &               & \\ \hline
  ~\cite{alrashdi2019ad}      & ExtraTrees, Random Forest            & Collaborative & Data \\ \hline
~\cite{garg2020multi} &
  \begin{tabular}[t]{@{}l@{}}Boruta, Firefly, Davies-Bouldin index-based $k$-medoid, DBSCAN,\\kernel-based locality sensitive hashing, $k$-distance graphs\end{tabular} &
  Collaborative &
  Data \\ \hline
~\cite{hasan2019attack}     & Random forest                        &               & \\ \hline
~\cite{markiewicz2020clust} & OPTICS (Ordering Points to Identify the Clustering Structure)                              & Collaborative & Data \\ \hline
~\cite{mergendahl2020rapid} & \begin{tabular}[t]{@{}l@{}}ARIMA (Autoregressive Integrated Moving Average), MLP (Multilayer Perceptron),\\LSTM (Long Short-Term Memory)\end{tabular}
                     & Collaborative & Data \\ \hline
~\cite{nguyen2019diot}      & GRU (Gated Recurrent Unit)                                & Federated     & Weights \\ \hline
~\cite{nie2020intrusion}    & $k$-means, mean shift, random forest &               & \\ \hline
~\cite{regan2022federated} & AE                                   & Federated & Weights \\  \hline
~\cite{shahid2019anomalous} & AE                                   & Collaborative & Data \\  \hline
~\cite{sriram2020network}   & DNN (Deep Neural Network)                               & Collaborative & Data \\ \hline
~\cite{wani2020ddos}        & MCOD (Multi-Level Outlier Detection), MLP                            & Collaborative & Data \\ \hline
~\cite{yamauchi2020anomaly} & Non-ML                               & Collaborative & Data \\ \hline
~\cite{yang2020ddos}        & AE, Decision Tree, PCA (Principal Component Analysis), IF, OCSVM                                   &               & \\
\hline
~\cite{campos2021evaluating}      & Multinomial logistic regression                                  & Federated     & Weights \\ \hline
~\cite{zhang2021federated}        & AE                                   & Federated & Weights\\ \hhline{====}
This paper & AE, $k$-means & Collaborative & Data \\ \hhline{====}
\end{tabular}%
}
\end{table*}

\subsection{Algorithmic complexity and flow}\label{subsec:algorithmiccomplexity_and_flow}

In multiple studies on IoT attack detection (e.g.,~\cite{meidan2018nbaiot,shahid2019anomalous,yang2020ddos}), only one algorithm was utilized. In contrast, we suggest a two-step method, where within each step a different algorithm is used (as further described and evaluated in Sections~\ref{sec:proposed_method} and~\ref{sec:evaluation_method}):
\begin{enumerate}[leftmargin=*]
    \item AE - to discern between frequent (`benign') and infrequent (possibly `malicious') traffic
    \item $k$-means - to further analyze only the infrequent traffic and discern between known (`rare yet benign') and unknown (`malicious') traffic to reduce the FPR
\end{enumerate}

Considering the existing methods (see Table~\ref{tab:literature_algorithms_collaboration}) that combine more than one algorithm, the research presented in~\cite{nie2020intrusion} suggests feature extraction using clustering analysis, with a combination of $k$-means and mean shift, followed by a random forest classifier, to detect attacks. Another method~\cite{pahl2018all} combines two clustering techniques: First, based on the packet interarrival time (IAT), clusters are built using $k$-means, and the abnormality value is calculated by BIRCH clustering. As opposed to our method, which was evaluated using flow-level data from multiple real IoT devices connected to real smart homes, this method was evaluated using emulated data from seven virtual state layer (VSL) services and three emulated IoT sites. Clustering was also employed in a more recent IoT-related study~\cite{dupont2021similarity}, however in that research the aim was to perform fingerprint-based classification rather than anomaly detection for attack detection.

\begin{table*}[ht]
\centering
\caption{Datasets used in prior research on anomaly/attack detection, mainly in IoT settings}
\label{tab:literature_datasets}
\resizebox{\textwidth}{!}{%
\begin{tabular}{c|l|l|l|l|r|r|l|l|r|c}
\hhline{===========}
\textit{\textbf{Ref.}} &
  \multicolumn{1}{c|}{\begin{tabular}[t]{@{}c@{}}\textit{\textbf{Type}}\\\textit{\textbf{of data}}\end{tabular}} &
  \multicolumn{1}{c|}{\begin{tabular}[t]{@{}c@{}}\textit{\textbf{Capturing}}\\\textit{\textbf{environment}}\end{tabular}} &
  \multicolumn{1}{c|}{\begin{tabular}[t]{@{}c@{}}\textit{\textbf{Number}}\\\textit{\textbf{ of devices/models}}\end{tabular}} &
  \multicolumn{1}{c|}{\begin{tabular}[t]{@{}c@{}}\textit{\textbf{Granularity}}\\\textit{\textbf{of instances}}\end{tabular}} &
  \multicolumn{1}{c|}{\begin{tabular}[t]{@{}c@{}}\textit{\textbf{Number}}\\\textit{\textbf{of instances}}\end{tabular}} &
  \multicolumn{1}{c|}{\begin{tabular}[t]{@{}c@{}}\textit{\textbf{Number}}\\\textit{\textbf{of features}}\end{tabular}} &
  \multicolumn{1}{c|}{\textit{\textbf{Class labels}}} &
  \multicolumn{1}{c|}{\begin{tabular}[t]{@{}c@{}}\textit{\textbf{Period}}\\\textit{\textbf{of time}}\end{tabular}} &
  \multicolumn{1}{c|}{\begin{tabular}[t]{@{}c@{}}\textit{\textbf{Size}}\\\textit{\textbf{(GB)}}\end{tabular}} &
  \textit{\textbf{Public}} \\ \hhline{===========}
~\cite{meidan2018nbaiot} &
  Generated &
  Laboratory &
  9 IoT devices &
  Packet &
  555,932 &
  115 &
  \begin{tabular}[t]{@{}l@{}}Benign/attacks related\\ to Mirai and BASHLITE\end{tabular}  &
  1 week &
   & \Checkmark
   \\ \hline
~\cite{pahl2018all} &
  Synthetic &
  Virtual state layer &
  \begin{tabular}[t]{@{}l@{}}7 VSL services,\\3 emulated IoT sites\end{tabular} &
  Packet &
  357,952 &
  13 &
  \begin{tabular}[t]{@{}l@{}}Normal, DoS, data type\\probing, malicious control,\\malicious operation,\\scan, spying, wrong setup\end{tabular}
   &
  24 hours &
  0.062 &
  \Checkmark \\ \hline
~\cite{alrashdi2019ad} &
  Generated &
  Laboratory &
   &
  Packet &
  2,540,044 &
  48 &
  Normal/attack &
  ~1 hour &
  100 &
  \Checkmark \\ \hline
\multirow{3}{*}{~\cite{garg2020multi}} & Generated &
  Audit logs (DARPA) &
   &
  Flow & 5,000,002
   & 22
   &
  DoS, privilage esccalation, probing &
  7 weeks &
  0.003 &
  \Checkmark \\ \cline{2-11} 
 &
  Real &
  University honeypots &
   & Packet
   & 3,046,267
   &
  24 &
  Attack or not & 2 months
   & 0.09
   &
  \Checkmark \\ \cline{2-11} 
 &
  \begin{tabular}[t]{@{}l@{}}Real,\\synthetic\end{tabular} &
  Yahoo (S5) &
   & Time series
   & 2,046,962
   &
  3 &
  Outliers and change-points &
   &
  0.016 &
  \Checkmark \\ \hline
~\cite{hasan2019attack} &
  Generated &
  Virtual state layer &
  \begin{tabular}[t]{@{}l@{}}7 VSL services,\\3 emulated IoT sites\end{tabular} &
  Packet &
  357,952 &
  13 &
  \begin{tabular}[t]{@{}l@{}}Normal, DoS, data type\\probing, malicious control,\\malicious operation,\\scan, spying, wrong setup\end{tabular} &
  24 hours &
  0.062 &
  \Checkmark \\ \hline
\multirow{2}{*}{~\cite{markiewicz2020clust}} &
  Generated &
  Laboratory &
  30 IoT devices &
  Packet &
   &
  9 & Benign only
   &
  20 days &
  7.3 &
  \Checkmark \\ \cline{2-11} 
 &
  Generated &
  Laboratory &
  5 IoT simulated scenarios &
  Packet &
  72,000,000 &
  14 &
  Normal/attack &
  5 scenarios &
  16.7 &
  \Checkmark \\ \hline
\multirow{6}{*}{~\cite{mergendahl2020rapid}} &
  Real &
  Smart home &
  9 IoT devices &
  Time window &
  500,000 &
  23 &
  Benign/anomalous &
   &
  0.002 &
  \Checkmark \\ \cline{2-11} 
 &
  Real &
  Smart home &
  28 IoT devices &
  Packet &
   &
  300 &
  DDoS attack &
  6 months &
  5.110 &
  \Checkmark \\ \cline{2-11} 
 &
  Real &
  Smart hospital &
  4 medical IoT devices &
   &
   &
   &
  DDoS attack &
  5 minutes &
  0.025 &
  \Checkmark \\ \cline{2-11} 
 &
  Real &
  Network telescope &
  1 network telescope &
   &
   &
   &
  DDoS attack traces &
  &
  656.6 &
  \Checkmark \\ \cline{2-11} 
 &
  Real &
  DDoS-as-a-service &
  7 booters &
   &
   &
   &
  DDoS attack traces &
   &
  250 &
  \Checkmark \\ \cline{2-11} 
 &
  Real &
  University &
   &
   &
   &
   &
  DDoS attack traces &
   &
  1 &
  \Checkmark \\ \hline
\multirow{3}{*}{~\cite{nguyen2019diot}} &
  Generated &
  Laboratory &
  33 IoT devices &
  Packet &
  2,087,280 &
  7 &
  Anomalous Y/N &
  165 hours &
  0.488 &
   \\ \cline{2-11} 
 &
  Real &
  Smart homes &
  14 IoT devices &
  Packet &
  2,286,697 &
  7 &
  Anomalous Y/N &
  2,352 hours &
  0.606 &
   \\ \cline{2-11} 
 &
  Generated &
  Laboratory &
  5 IoT devices &
  Packet &
  21,919,273 &
  7 &
  Anomalous Y/N &
  84 hours &
  8.110 &
   \\ \hline
~\cite{nie2020intrusion} &
  Simulated &
  Laboratory (ISCX 2012) &
   Devices in a testbed & Flow
   & 1,526,148
   &
   &
  Normal/malicious &
  7 days &
  84.42 &
  \Checkmark \\ \hline
~\cite{shahid2019anomalous} &
  Real &
  Smart home &
  4 IoT devices &
  TCP flow &
  46,796 &
  16 &
  Benign/anomalous &
  1 day &
   &
   \\ \hline
\multirow{2}{*}{~\cite{sriram2020network}} &
  Real &
  Smart home &
  9 IoT devices &
  Time window &
  500,000 &
  23 &
  Benign/anomalous &
   &
  0.002 &
  \Checkmark \\ \cline{2-11} 
 &
  Generated &
  Laboratory &
  5 IoT simulated scenarios &
  Packet &
  72,000,000 &
  14 &
  Normal/attack &
  5 scenarios &
  16.7 &
  \Checkmark \\ \hline
~\cite{wani2020ddos} &
  Synthetic &
  Laboratory & 1 simulation tool
   & Flow
   & 1,054
   & 7
   &
  Benign/malicious &
   &
   &
   \\ \hline
~\cite{yamauchi2020anomaly} &
  Generated &
  Laboratory &
  28 IoT devices & Sensor data
   & 
   &
   &
  Anomalous Y/N &
  6 months &
   &
   \\ \hline
\multirow{3}{*}{~\cite{yang2020ddos}} &
  Synthetic &
  Simulation (SYNT) & 1 simulation tool
   &
  Packet &
  152,230 &
  27 &
  Normal/attack &
   &
   &
   \\ \cline{2-11} 
 &
  Generated & Lab (UNB)
   & 12 computers
   & Packet
   & 95,785
   & 27
   &
  Normal/attack &
  5 days &
  51.1 &
  \Checkmark \\ \cline{2-11} 
 &
  Real & US-Japan lines (MAWI)
   & 5 sample points
   & Packet
   & 62,000
   & 27
   &
  Normal & 12 months
   & 
   &
  \Checkmark \\ 
  \hhline{===========}
\begin{tabular}[t]{@{}c@{}}This\\ paper\end{tabular} &
  \begin{tabular}[t]{@{}l@{}}Real,\\generated\end{tabular} &
  \begin{tabular}[t]{@{}l@{}}5 Smart homes,\\laboratory\end{tabular} &
  8 identical IoT devices &
  IPFIX &
  200,087 &
  24 &
  \begin{tabular}[t]{@{}l@{}}`Assumed benign,' 'rare yet\\benign,' 'malicious' (crypto-\\mining, Nmap scanning)\end{tabular} &
  21 days &
  \multicolumn{1}{r|}{0.0178} &
  \Checkmark \\ \hhline{===========}
\end{tabular}%
}
\end{table*}

\subsection{Collaborative anomaly detection}\label{subsec:collaborative_anomaly_detection}

Although in many cases~\cite{meidan2018nbaiot,pahl2018all,wani2020ddos}, traffic-based IoT anomaly detection methods demonstrate excellent performance in detecting cyber-attacks (i.e., high recall), such methods also tend to suffer from a non-negligible FPR. One reason for this commonly seen drawback is that traffic anomalies can also be generated for non-malicious activities, such as infrequent (yet benign) user activities and contextual changes. Although infrequent, such actions should not trigger unnecessary risk-mitigating countermeasures for allegedly-detected cyber-attacks. To address this issue and reduce the FPR, we propose training the anomaly detector in a \emph{collaborative} manner via data sharing while enriching the data with collaboratively engineered features. As part of our method, traffic data is collected from multiple IoT devices of the same model (i.e., many identical instances of an IoT product (of the same version), such as streamer.Amazon.Fire\_TV\_Gen\_3~\cite{firetv}), which are connected to various home networks and operated by different people. The motivation for training the anomaly detector in a collaborative manner using shared data is that the inherent richness and heterogeneity of such data allows the capturing of enough examples of similar \emph{local} abnormalities which are rare yet benign, whereby \emph{globally} they would be identified by the anomaly detector as sufficiently-frequent normal patterns.

Collaborative intrusion detection systems (CIDSs) are network intrusion detection systems (NIDSs) whose data is collected from various monitors that function as sensors (disparate IoT devices in our use case); these systems are comprised of several units that perform intrusion detection on the data collected from the sensors~\cite{vasilomanolakis2015taxonomy}. There are a few main CIDS architectures. The first is referred to as a \emph{centralized} CIDS (data sharing) in which the data is sent to a central unit where the model performs computations on the data. Another architecture is referred to as a \emph{decentralized} CIDS where there are several computation units located between the sensors and the model that deal with data aggregation and preprocessing. \emph{Distributed} CIDSs are a form of collaboration architecture in which there is no central component and the analysis is done within the data monitors.

Many of the prior studies on (IoT-related) CIDSs are listed in Table~\ref{tab:literature_algorithms_collaboration}. The authors of~\cite{markiewicz2020clust} based their method on the OPTICS algorithm for clustering (and cluster membership), and evaluated their method using IoT datasets to detect malicious traffic. In~\cite{shahid2019anomalous}, the authors presented a CIDS with multiple sparse AEs, one for each IoT device type. To evaluate their method, they used a dataset from one smart home with four different devices, however they did not publish their dataset, so their results cannot be reproduced. Two additional CIDSs, in which the proposed method is multi-staged (similar to our approach), are~\cite{pahl2018all} and~\cite{garg2020multi}. The former combines ARIMA, MLP, and LSTM to detect DDoS attacks, while the authors of the latter proposed a multi-stage CIDS which also combines active learning, however the experiments in both of these studies were conducted using non-IoT datasets.

In this research, similar to some of the studies presented in Table~\ref{tab:literature_algorithms_collaboration}, we chose to implement the CIDS via data sharing, using a centralized architecture. The main reason for this choice is that in our use case (see Subsection~\ref{subsec:assumptions}), the traffic data of the monitored IoT devices is \emph{already} collected in a central cloud server, and then it is preprocessed and analyzed in another existing server, incurring little overhead. Moreover, we do not assume a minimal level of processing power of the monitored IoT devices or home routers, or the users' consent to access their devices and use them for storage and/or computation; therefore, we prefer not to rely on the monitored IoT devices but rather on the readily-available central servers. As acknowledged by the authors of~\cite{Wu2018ASystems}, in reality, the communication and computational resources of the end nodes are typically limited and there may also be a time delay.

Given our preference for collaboration in the anomaly detection model's training, we could have opted for federated learning (FL)~\cite{konevcny2016federated,mcmahan2017communication}, which was previously used for IoT security purposes~\cite{nguyen2019diot}. As part of the FL collaborative approach, the sensors train (local) models and share their parameters (e.g., AE weights~\cite{regan2022federated}) instead of sharing traffic data, mainly for privacy preservation and to limit the overhead of communication and storage~\cite{zhang2021federated}. However, in real-world scenarios, the deployment of FL on (typically resource-constrained) IoT devices might be infeasible due to memory, computing power, and energy consumption limitations~\cite{campos2021evaluating}; such FL deployment would also necessitate obtaining the users' consent to remotely access their IoT devices to install the security software, which may also be infeasible. Moreover, FL often introduces a trade-off between privacy preservation and model performance in terms of accuracy~\cite{liu2020privacy}, efficiency~\cite{yang2019federated}, and/or convergence~\cite{wei2020federated}. Given the above, in this paper, as we wish to develop a feasible and high-performing CIDS mechanism for IoT device anomaly detection, we prefer the centralized collaborative approach over FL.

\subsection{Existing public datasets for IoT attack detection}\label{subsec:existing_public_datasets}

The availability of a scientific dataset is important for both \emph{(a)} reproducing and validating related research, and \emph{(b)} quantitatively evaluating future methods. In the NIDS domain, it was recently noted~\cite{mahfouz2020ensemble} that there is a dearth of datasets. This is also seen in the subdomain of IoT-related NIDSs, where there is a shortage of publicly available datasets, particularly high-quality datasets that \emph{(a)} contain data captured over a long period of time and are \emph{(b)} sufficiently representative of real-world settings, \emph{(c)} properly labeled with present-day attacks, \emph{(d)} comprised of data from multiple real devices and locations, such that they can be reliably used to evaluate novel methods for CIDSs. Table~\ref{tab:literature_datasets} provides a description of the datasets used in prior research on (IoT) anomaly and attack detection. As can be seen in the table, not all of the datasets have been made publicly available. Of those that are publicly available, in some cases the research addresses IoT anomaly or attack detection but uses non-IoT data for evaluation~\cite{garg2020multi}. The data of other datasets was captured in a period of just minutes~\cite{mergendahl2020rapid} or hours~\cite{hasan2019attack,pahl2018all}. Other studies used data which is simulated~\cite{nie2020intrusion}, synthetic~\cite{garg2020multi}, or generated in a lab~\cite{markiewicz2020clust,meidan2018nbaiot,sriram2020network}, sometimes using Raspberry PIs~\cite{zhang2021federated} instead of real IoT devices. Even if the data was captured from real-world settings, usually the dataset includes the data from just a small number of IoT devices~\cite{shahid2019anomalous} or a honeypot~\cite{garg2020multi}, preventing the proper evaluation of a CIDS. With regard to the empirical evaluation of collaborative and FL approaches, another major limitation of existing datasets is that in most cases, the separation into nodes is only superficial. That is, traffic data is collected from just one IoT device deployed in one location and is later split using various strategies~\cite{rey2022federated,regan2022federated}. These data splits mimic nodes that perform collaboration, however the data splits fail to represent the diversity of identical devices which are deployed separately and interacted with by different users.

In comparison, in this research we utilize (and make publicly available) a novel dataset whose data was collected from eight identical IoT devices, deployed in six disparate networks, five of which are \emph{real} home networks where the data is assumed benign, and one of which is in our controlled lab's network where we could implement IoT-related attacks and accurately label the traffic accordingly. In all of the homes the same IoT model was monitored, while different people in different surroundings and network settings freely interacted with the same IoT model. This allows us to quantitatively evaluate the proposed CADeSH method and set a benchmark for future (IoT-related, collaborative) NIDSs. In addition, our unique dataset covers a period of 21 days, which is relatively long time period compared to that of the datasets listed in Table~\ref{tab:literature_datasets}. Another important aspect of our dataset is that unlike most datasets used thus far (some of which were published), it is not comprised of packet-level raw traffic data but rather is comprised mainly of IPFIX-level aggregated \emph{meta}data, which results in lower communication/storage overhead while preserving privacy. We note that in~\cite{Pashamokhtari2021Inferring}, a large and diverse IPFIX-based dataset was collected from an IoT testbed, however it \emph{(1)} only contains benign data, limiting attack detection evaluation, \emph{(2)} emanates from a single network only, such that collaborative learning can not be properly evaluated, and \emph{(3)} reflects a limited frequency of human interaction, which is performed by just the lab staff, as opposed to the ongoing and natural real-world interaction with a diverse range of people in multiple disparate home networks reflected in our dataset.

Further details regarding the acquisition and properties of our novel dataset are provided in Section~\ref{sec:evaluation_method}.

\begin{figure}[ht]
\centering
\includegraphics[width=0.5\textwidth, trim={0 10.2cm 0 0cm},clip]{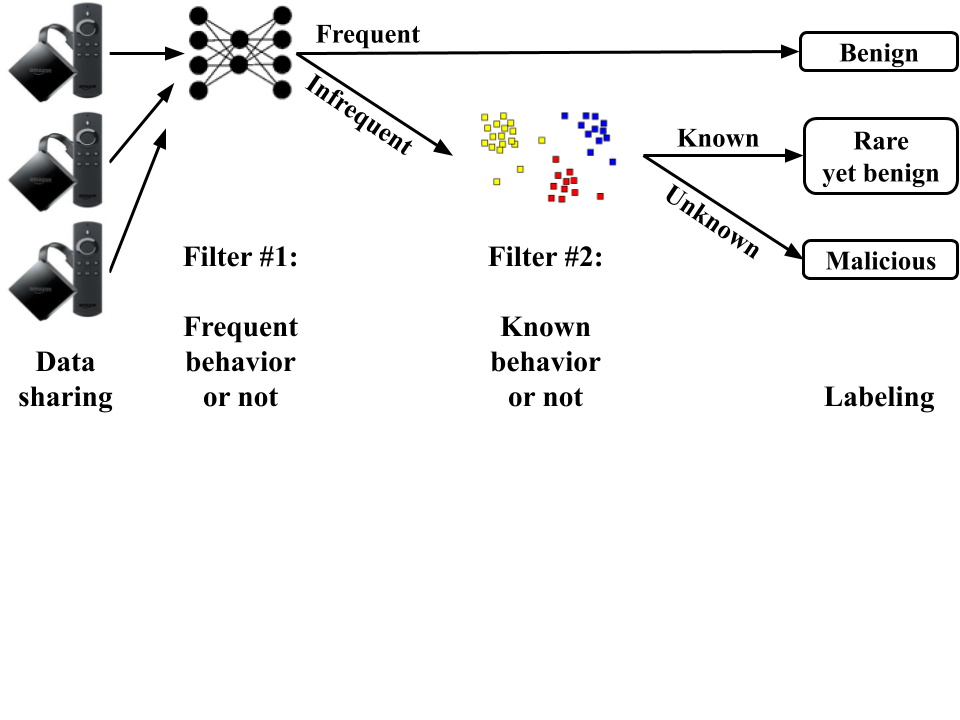}
\caption{Our two-step method for collaborative IoT anomaly detection}
\label{fig:anomaly_detection_algo}
\end{figure}

\section{Proposed method}\label{sec:proposed_method}

Overall, as illustrated in Fig.~\ref{fig:anomaly_detection_algo}, the method we propose for (collaborative) IoT anomaly/attack detection in smart homes consists of two main steps. We refer to each of those steps as a \textit{filter}, as some flows continue to the next step for further analysis while others are assigned a final label. Our method is trained (and is later used for inference) based on shared flow-level traffic data $F^m$ that is collected from multiple identical model $m$ IoT devices $D^m: \{ d^m_1, d^m_2,\dots ,d^m_n\}$, deployed in disparate home networks and interacted with by different users. As in any anomaly-based IDS, during training we rely on data which is assumed to be mostly benign so that a profile of the normal traffic patterns can be learned. To be on the safe side, our method includes also data selection, cleansing, and sanitization stages prior to training, as described in Subsection~\ref{subsec:Data Enrichment_preprocessing_and_Partitioning}. Later, during inference, each outbound traffic flow $f^m$ is first labeled by $Filter_1^m$ as frequent (i.e., `benign' in terms of attack detection) or infrequent (i.e., somewhat anomalous, but not necessarily `malicious'). Afterwards, if labeled as infrequent by $Filter_1^m$, the flow is further labeled by $Filter_2^m$ as known behavior (i.e., `rare yet benign') or unknown (i.e., highly anomalous, presumably `malicious').

To implement $Filter_1^m$ and $Filter_2^m$, we propose (and empirically evaluate, see Sections~\ref{sec:evaluation_method} and~\ref{sec:results}) the use of AEs and clusters, respectively. The reason for the former choice is that AEs have demonstrated effectiveness in binary IoT anomaly detection~\cite{meidan2018nbaiot,shahid2019anomalous,yang2020ddos,zhang2021federated} and therefore could also facilitate the distinction between frequent and infrequent behaviors. Regarding the latter choice, the intuition is that rare yet benign traffic data can be divided into multiple clusters which represent various infrequent activities (slight anomalies such as a webcam booting or zooming in/out, etc.). A traffic flow $f^m$ which is too far from any of the rare yet benign clusters is considered a strong anomaly, presumably representing a cyber-attack.

To describe our method in further detail, we use the following notation:
\begin{compactitem}[rightmargin=*]
\item [$m$:] An IoT model, defined~\cite{meidan2020deNAT} as the combination of type, manufacturer, and model number. For instance, \emph{webcam.D\_Link.CS\_930LB1}~\cite{dlinkcs930} and \emph{webcam.D\_Link.DCS\_933L}~\cite{dlinkcs933} are two distinct IoT models that share the same \emph{type} and \emph{manufacturer}.
\item [$D^m$:] A set of \(n\) identical IoT devices $\{ d^m_1, d^m_2, \dots ,d^m_n\}$ defined by their MAC addresses, all of which are instances of model \(m\). 
\item [$f_i^m$:] A single traffic flow that emerged from device \(d^m_i\), enriched with features from additional data sources such as DNS requests or the Webroot API; interchangeably referred to in this paper as an IPFIX, flow, record, or instance.
\item [$F_i^m$:] A set of \(x\) traffic flows $\{ f^m_{i1}, \dots ,f^m_{ix}\}$ that emerged from \(d^m_i\).
\item [$F^m$:] A superset of \(F^m_i\), containing flows from all of the devices in \(D^m\). We partition $F^m$ chronologically into three mutually exclusive parts: $F^m_{training}$ for training the filters, $F^m_{validation}$ for model optimization and anomaly threshold calibration, and $F^m_{test}$ for anomaly/attack detection evaluation.
\item [$Filter_1^m$:] The first anomaly detector, trained to determine whether a flow \(f^m_i\) represents a frequent or infrequent activity of device \(d^m_i\).
\item [$Filter_2^m$:] The second (subsequent) anomaly detector, trained to determine whether \(f^m_i\), which was already labeled by $Filter_1^m$ as infrequent, represents a known or unknown activity of device \(d^m_i\).
\item [$k^*$:] The optimal number of clusters in $Filter_2^m$, assuming $k$-means is used.
\item [$MSE^m_f$:] The mean squared error (i.e., the AE's reconstruction error), obtained by applying $Filter_1^m$ to a flow $f^m$. Since $Filter_1^m$ is trained using only assumed-benign data, a high $MSE^m_f$ value indicates a severe abnormality as the AE fails to reconstruct $f^m$.
\item [$epochs_{max}$:] A parameter to limit the number of epochs during which the AE's weights in $Filter_1^m$ are updated.
\item [$delta_{min}$:] A parameter to limit the duration of $Filter_1^m$'s training and decrease the chances that $Filter_1^m$ overfits the training data. $delta$ is the difference between the current and previous $MSE^m$ which is calculated using $F^m_{validation}$.
\item [$patience_{max}$:] An upper bound for the number of subsequent AE training epochs where $delta<delta_{min}$ (i.e., no substantial decrease in MSE is obtained). Reaching $patience_{max}$ is an indication of model convergence, and it is used to stop the AE's training even if $epochs_{max}$ has not been reached.
\item [$TH^m_{frequent}$:] Frequency threshold on $MSE^m$. Any (future or test) flow $f^m$ is labeled as infrequent if its reconstruction error $MSE_f^m$ exceeds $TH^m_{frequent}$, or as frequent otherwise. We propose setting $TH^m_{frequent}$ as a certain percentile (denoted as $pctl_{frequent}$) of $MSE^m$ on $F^m_{validation}$, e.g., the $60^{th}$ or $70^{th}$ percentile.
\item [$\hat{f^m_{frequent}}$:] The label assigned to a flow $f^m$ by $Filter_1^m$; it could be either frequent or infrequent. In our empirical evaluation, we denote the subsets of flows which were labeled infrequent by $Filter_1^m$ as $F^{m,\:infrq}_{training}$, $F^{m,\:infrq}_{validation}$, and $F^{m,\:infrq}_{test}$.
\item [$TH^m_{known}$:] A set of $k^*$ distance thresholds $th^m_{known}$, one for each cluster in $Filter_2^m$. Any (future or test) flow $f^m$ which was labeled as infrequent by $Filter_1^m$ is labeled by $Filter_2^m$ as unknown if its distance $d^{f,\:cluster}$ from the trained cluster to which it is assigned exceeds the related $th^m_{known}$ or as rare yet benign otherwise. We propose setting the cluster-specific thresholds in $TH^m_{known}$ as a certain percentile (denoted as $pctl_{known}$) of the related cluster distances $D^{cluster}$ on $F^{m,\:infrq}_{validation}$, e.g., the $95^{th}$, $98^{th}$, or $100^{th}$ percentile.
\item [$\hat{f^m_{known}}$:] The label assigned to a flow $f^m$ by $Filter_2^m$; could be either known (`rare yet benign') or unknown (`malicious').
\end{compactitem}

\subsection{$Filter_1^m$: frequent behavior or not}\label{subsec:filter_1_frequent}

For a given IoT model $m$, the first step of CADeSH handles the training, optimization, and application of an anomaly detector $Filter^m_1$, which labels each outbound flow $f^m_i$ from an IoT device $d^m_i$ as either frequent or infrequent. As described in Algorithm~\ref{alg:train_AE}, $Filter^m_1$ is trained and optimized using $F^m_{training}$ and $F^m_{validation}$, respectively, which are collected from multiple identical IoT devices $D^m$ deployed in multiple domestic networks and thus are regarded as highly representative of the typical traffic data associated with $m$. Since we do not have the actual ground truth label (i.e., `benign' or `malicious') for each flow $f^m_i$ gathered from outside our controlled lab, we assume that most of the flows in $F^m$ are benign and represent normal activity of the IoT devices of the respective model $m$. To increase the probability of using benign flows, we perform data sanitization (see Subsection~\ref{subsec:Data Enrichment_preprocessing_and_Partitioning}). The weights of the AE underlying $Filter^m_1$ are updated using $F^m_{training}$ with MSE serving as the loss function. Training is stopped once the number of epochs exceeds $epochs_{max}$ or when the conditions for early stopping~\cite{keras_early_stopping} are met, namely a sequence of epochs without a significant decrease of the MSE.

{\small
\begin{algorithm}[!h]
  \footnotesize
\hspace*{\algorithmicindent} \textbf{Input:} $F^m_{training}$, $F^m_{validation}$, $epochs_{max}$, $delta_{min}$, $patience_{max}$\\
 \hspace*{\algorithmicindent} \textbf{Output:} $Filter^m_{1}$
\begin{algorithmic}[1]
\State $epoch \gets 1$
\State $epoch\_low\_MSE \gets 1$  
\State $delta \gets \infty$
\State $MSE_{previous} \gets 0$ 
\While{$epoch < epochs_{max}$ \& $delta > delta_{min}$ \& $epoch\_low\_MSE<patience_{max}$} 
    \State $Filter^m_{1} \gets filter^m_{1}.update\_weights(F^m_{training})$
  	\State $MSE_{current} \gets filter^m_{1}.calc.\_MSE(F^m_{validation})$
   	\State $delta \gets MSE_{current} - MSE_{previous}$
  	\State $MSE_{previous} \gets MSE_{current}$
  	\State $epoch \gets epoch + 1$
  	\If{$delta<delta_{min}$}
  	    \State $epoch\_low\_MSE \gets epoch\_low\_MSE+1$
  	\EndIf
\EndWhile
\State \Return $Filter^m_{1}$
\end{algorithmic}
\caption{Train the AE-based $Filter^m_1$}\label{alg:train_AE}
\end{algorithm}
}

{\small
\begin{algorithm}[ht]
  \footnotesize
\hspace*{\algorithmicindent} \textbf{Input:} $Filter^m_{1}$, $F^m_{validation}$, $pctl_{frequent}$ \\
 \hspace*{\algorithmicindent} \textbf{Output:} $TH^m_{frequent}$
\begin{algorithmic}[1]
\State $MSE\_{validation} \gets []$
\For{$f^m$ in $F^m_{validation}$}
    \State $MSE^m_f \gets filter^m_{1}.calc.\_MSE(f^m)$
    \State $MSE\_{validation} \gets MSE\_{validation}.append(MSE^m_f)$
\EndFor
\State $MSE\_{validation} \gets MSE\_{validation}.sort()$
\State $TH^m_{frequent} \gets MSE\_{validation}.calc.\_percentile(pctl_{frequent})$
\State \Return $TH^m_{frequent}$
\end{algorithmic}
\caption{Set a frequency threshold for $Filter_1^m$}\label{alg:set_frequency_threshold}
\end{algorithm}
}

Algorithm~\ref{alg:train_AE} is used to train an anomaly detector $Filter^m_1$ which captures assumed-benign activity of model $m$'s IoT devices. After $Filter^m_1$ has been trained, a decision regarding how to label each future or test $f^m$ ('frequent' or `infrequent') must be made. In order to set the frequency threshold $TH^m_{frequent}$, as described in Algorithm~\ref{alg:set_frequency_threshold}, we apply the AE-based $Filter^m_1$ to $F^m_{validation}$ and observe the distribution of the MSE. The higher the MSE value, the less frequent (more abnormal) $f^m$ is considered to be. We set the value of $TH^m_{frequent}$ as a certain percentile of $MSE^m_f$'s distribution on $F^m_{validation}$, e.g., the $60^{th}$ or $70^{th}$ percentile. This number reflects our expectation, based on findings from prior research~\cite{bertino2017botnets,cvitic2020definition,Doshi2018MachineDevices,meidan2018nbaiot,Skowron2020}, that for IoT devices, a large portion of the normal traffic data is relatively repetitive, stable, and predictable, and thus its MSE should be rather low.

Eventually, as described in Algorithm~\ref{alg:determine_frequent_or_not}, the label $\hat{f^m_{frequent}}$ of a given test flow is determined by comparing its $MSE^m_f$ (calculated using $Filter^m_1$) with the threshold $th^m_{frequent}$. A flow whose $MSE^m_f<th^m_{frequent}$ is labeled as frequent (i.e., no suspicion of any malicious activity), while a flow whose $MSE^m_f \ge th^m_{frequent}$ is temporarily labeled as infrequent and then inspected by $Filter^m_2$ to assign a final label, either known (rare yet benign) or unknown (malicious).

{\small
\begin{algorithm}[!h]
  \footnotesize
\hspace*{\algorithmicindent} \textbf{Input:} $f^m$, $Filter^m_{1}$, $th^m_{frequent}$ \\
 \hspace*{\algorithmicindent} \textbf{Output:} $\hat{f^m_{frequent}}$
\begin{algorithmic}[1]
    \State $MSE^m_f \gets filter^m_{1}.calc.\_MSE(f^m)$
    \If{$MSE^m_f < th^m_{frequent}$} \State $\hat{f^m_{frequent}} \gets True$
    \Else \State $\hat{f^m_{frequent}} \gets False$
    \EndIf
\State \Return $\hat{f^m_{frequent}}$
\end{algorithmic}
\caption{Determine whether a flow is frequent or not using $Filter_1^m$}\label{alg:determine_frequent_or_not}
\end{algorithm}
}

\subsection{$Filter_2^m$: known behavior or not}\label{subsec:filter_2_known}

This component of our method \emph{only} handles the flows that were previously classified by $Filter_1^m$ as infrequent and determines whether their abnormality is extreme enough to be considered a completely unknown behavior (presumably indicating a malicious activity) or is a known/legitimate behavior, i.e., rare yet benign. In this component, we propose (and empirically evaluate) the use of a clustering algorithm such as $k$-means~\cite{nie2020intrusion,pahl2018all}, assuming that the trained clusters in $Filter_2^m$ can represent various rare yet benign traffic patterns of an IoT model $m$.

{\small
\begin{algorithm}[h]
  \footnotesize
\hspace*{\algorithmicindent} \textbf{Input:} $F^{m,\:infrq}_{training}$, $k_{min}$, $k_{max}$\\
 \hspace*{\algorithmicindent} \textbf{Output:} $Filter^m_{2}$
\begin{algorithmic}[1]
\State $Avg\_Sil.\_Scores\gets []$
\For{$k_{temp}$ in range($k_{min}$,$\:k_{max}$)}
    \State $clust.\_assignments_{temp} \gets k\_means.fit\_predict(F^{m,\:infrq}_{training},\:k_{temp})$
    \State $Avg\_Sil.\_Score_{temp} \gets calc\_Avg\_Sil.\_Score(clust.\_assignments_{temp})$
    \State $Avg\_Sil.\_Scores \gets Avg\_Sil.\_Scores.append(Avg\_Sil.\_Score_{temp})$
\EndFor
\State $k^* \gets arg\_max(Avg\_Sil.\_Scores)$
\State $Filter^m_{2} \gets k\_means.fit(F^{m,\:infrq}_{training},\:k^*)$
\State \Return $Filter^m_{2}$
\end{algorithmic}
\caption{Train the ($k$-means) clustering-based $Filter^m_2$}\label{alg2:train_filter_2}
\end{algorithm}
}

To train and optimize the clustering model which implements $Filter_2^m$, we use $F^{m,\:infrq}_{training}$, which is a subset of $F^m_{training}$ where $MSE^m \ge TH^m_{frequent}$ (i.e., training instances previously classified as infrequent by $Filter_1^m$). This allows the model to focus only on data which is not entirely normal, such that better separation can be learned between real (malicious) anomalies and mediocre (rare yet benign) anomalies. As outlined in Algorithm~\ref{alg2:train_filter_2}, we find the optimal number of clusters $k^*$ by training a $k$-means model for each $k$ from the range $[k_{min}, ..., k_{max}]$ and selecting $k$ which maximizes the average silhouette score~\cite{shahapure2020cluster}.  

{\small
\begin{algorithm}[!h]
  \footnotesize
\hspace*{\algorithmicindent} \textbf{Input:} $Filter^m_{2}$, $F^{m,\:infrq}_{validation}$, $pctl_{known}$ \\
 \hspace*{\algorithmicindent} \textbf{Output:} $TH^m_{known}$
\begin{algorithmic}[1]
\State $TH^m_{known} \gets []$
\For{$cluster$ in $Filter^m_{2}$}
    \State $D^{cluster} \gets []$
    \For{$f$ in $F^{m,\:infrq}_{validation}$}
    \If{$f \in cluster$} \State $d^{f,\:cluster} \gets distance(f,\:cluster_{center})$
    \State $D^{cluster} \gets D^{cluster}.append(d^{f,\:cluster})$
    \EndIf
\EndFor
    \State $D^{cluster} \gets D^{cluster}.sort()$
    \State $th^{m,\:cluster}_{known} \gets D^{cluster}.calc.\_percentile(pctl_{known})$
    \State $TH^m_{known} \gets TH^m_{known}.append(th^m_{known})$
\EndFor
\State \Return $TH^m_{known}$
\end{algorithmic}
\caption{Set distance thresholds for $Filter_2^m$}\label{alg:set_frequency_threshold_f2}
\end{algorithm}
}

\begin{table*}[!t]
\centering
\caption{Number of monitoring days, device locations, monitored devices, and IPFIXs for each label across our data partitions}
\label{tab:daily_labels}
\resizebox{1.0\textwidth}{!}{
\bgroup
\def\arraystretch{1.3}
\begin{tabular}{l|c|l|c|rrrr}
\hhline{========}
\multicolumn{1}{c|}{\textbf{Data}}      & \textbf{Number of}       & \multicolumn{1}{c|}{\textbf{Device}}      & \textbf{Number of}         & \multicolumn{4}{c}{\textbf{Number of IPFIXs}}                                                                                                                                                     \\ \cline{5-8} 
\multicolumn{1}{c|}{\textbf{partition}} & \textbf{monitoring days} & \multicolumn{1}{c|}{\textbf{location(s)}} & \textbf{monitored devices} & \multicolumn{1}{c|}{\textbf{Total}} & \multicolumn{1}{c|}{\textbf{Assumed benign}} & \multicolumn{1}{c|}{\textbf{Being scanned by Nmap}} & \multicolumn{1}{c}{\textbf{Is executing cryptomining}} \\ \hhline{========}
$F^m_{training}$                                & 13                       & \multirow{2}{*}{5 X Non-lab}                  & \multirow{2}{*}{7}         & \multicolumn{1}{r|}{133,351}       & \multicolumn{1}{r|}{133,351}                & \multicolumn{1}{r|}{}                               &                                                        \\ \cline{1-2} \cline{5-8} 
$F^m_{validation}$                              & 3                        &                                           &                            & \multicolumn{1}{r|}{39,481}         & \multicolumn{1}{r|}{39,481}                  & \multicolumn{1}{r|}{}                               &                                                        \\ \hline
$F^m_{test}$                                    & 5                        & 1 X Lab only                                  & 1                          & \multicolumn{1}{r|}{27,255}         & \multicolumn{1}{r|}{22,472}                  & \multicolumn{1}{r|}{3,080}                          & 1,703                                                  \\ \hhline{========}
\end{tabular}
\egroup
}
\end{table*}

Then, to set a maximum distance threshold $TH^m_{known}$ for each of the trained $k^*$ clusters, beyond which a (future or test) instance is considered abnormal enough to be labeled as unknown (malicious), we use Algorithm~\ref{alg:set_frequency_threshold_f2}: We assign each (infrequent though assumed-benign) instance $f$ from $F^{m,\:infrq}_{validation}$ to its nearest cluster using $Filter_2^m$ and observe the distribution of distances $D^{cluster}$ between each instance and the center of its assigned cluster. For each cluster, the abnormality threshold $th^m_{known}$ is set to be a certain percentile of that cluster's distance distribution, e.g., the $99^{th}$ or $95^{th}$ percentile of $D^{cluster}$. The rationale for doing so is that the distribution of cluster distances on $F^{m,\:infrq}_{validation}$ should be representative of infrequent though assumed-benign data, so that extremely high distances ought to raise suspicion concerning malicious activities. 

{\small
\begin{algorithm}[!h]
  \footnotesize
\hspace*{\algorithmicindent} \textbf{Input:} $f$, $Filter^m_{2}$, $th^m_{known}$ \\
 \hspace*{\algorithmicindent} \textbf{Output:} $\hat{f^m_{known}}$
\begin{algorithmic}[1]
\State $cluster \gets arg\_min(distance(f,\:cluster_{center}))$
\State $d^{f,\:cluster} \gets distance(f,\:cluster_{center})$
    \If{$d^{f,\:cluster} < th^m_{known}$} \State $\hat{f^m_{known}} \gets True$
    \Else \State $\hat{f^m_{known}} \gets False$
    \EndIf
\State \Return $\hat{f^m_{known}}$
\end{algorithmic}
\caption{Determine whether a flow is known or not using $Filter_2^m$}\label{alg:determine_known_or_not}
\end{algorithm}
}
Eventually, as outlined in Algorithm~\ref{alg:determine_known_or_not}, a final decision for infrequent test instances is made based on \emph{(1)} assigning them to the nearest cluster and \emph{(2)} comparing them to their cluster-specific distance threshold $th^m_{known}$.

\section{Evaluation method}\label{sec:evaluation_method}

\subsection{Experimental setup}\label{subsec:experimental_setup}

\begin{figure}[!h]
\centering
\includegraphics[width=0.5\textwidth, trim={0 0 0 2.3cm},clip]{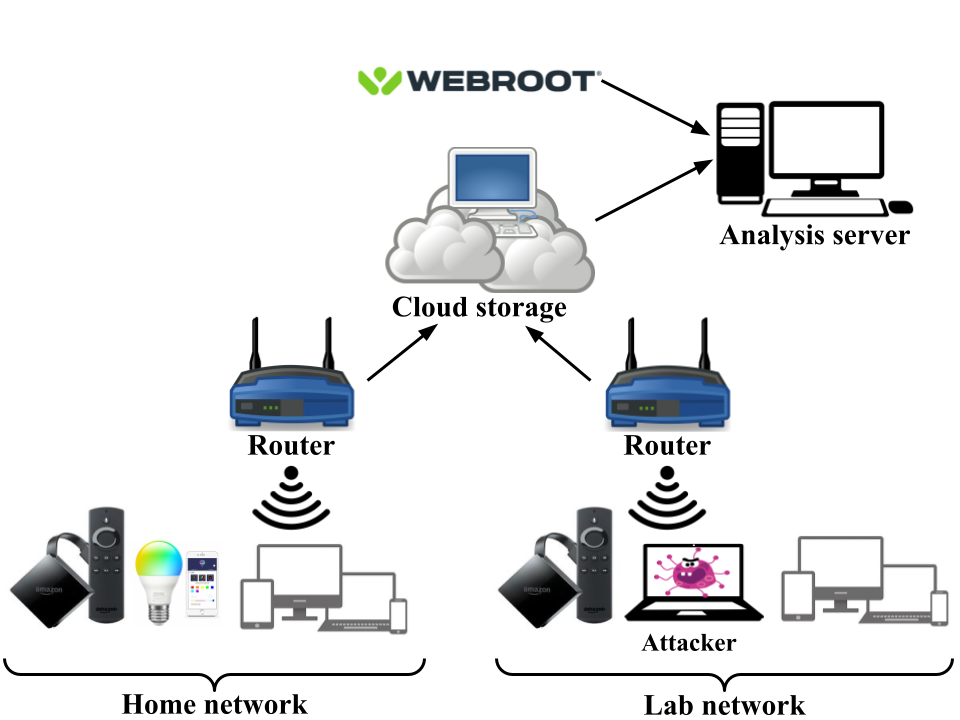}
\caption{Our experimental setup}
\label{fig:setup}
\end{figure}

We were motivated to conduct this research by a company that provides cyber-defense services for smart homes, and this company provided the data used in our experiments. To protect the company's subscribers against various IoT-related attacks, flow-level traffic data is continuously captured from the monitored home networks, stored in a cloud server, and then augmented and analyzed centrally to enable risk-mitigating responses. As can be seen in Fig.~\ref{fig:setup}, in each home network various devices (both IoT and non-IoT devices) are connected to the Internet via a router. This router sends the traffic data to a designated cloud storage server, from which data is sent to another server for analysis. Note that our lab network has the exact same setup as the home networks. The only difference is that in our lab we connected a laptop (denoted as `Attacker' in Fig.~\ref{fig:setup}), which was used to implement cyber-attacks, to the network (see Subsection~\ref{subsec:implemented_attacks_and_Corresponding_Labels}).

The abovementioned cyber-defense company kindly shared the following data with us: network traffic flows which were collected during a period of 21 days (see Table~\ref{tab:daily_labels}), plus the related DNS requests and responses and reputation intelligence of the destination IP addresses, produced by Webroot. To evaluate our IoT anomaly/attack detection method, we implemented the cyber-attacks during that 21 day period, such that the company's network traffic flows were collected at the same time, using the same setup as the other (home) networks. In our lab, as described next, we merged the various data sources, added a number of engineered features, and formed a single dataset for experimentation. For this process of data merging, engineering, preprocessing, and analysis, we used a Windows 2016 server with a 2GHz E5-2620 CPU and 256GB RAM.

\subsection{Dataset}\label{subsec:dataset}

\textbf{Instances.} Each instance in our dataset represents an outbound network traffic flow (in the form of an IPFIX~\cite{ipfixIETF,hofstede2014flowIPFIX,cisconetflowipfix}), enriched using additional data sources. An IPFIX is a 5-tuple aggregation of the raw traffic data (source IP address, destination IP address, source port, destination port, and IP protocol). Most of the IPFIXs in our dataset were collected from real home networks by the company we cooperated with, and the remaining flows were collected in our lab, using the same setup used by the company's home subscribers. In our experiments, we concentrated on the IoT model, streamer.Amazon.Fire\_TV\_Gen\_3~\cite{firetv}, because it was deployed in five monitored home networks and we could empirically evaluate our collaborative algorithm and its generalizability to a sixth network (in our lab).

\textbf{Features.}
Each instance is comprised of several feature groups (Appendix~\ref{appendix:features} contains a list of the feature names in each feature group, as well as brief descriptions):
\begin{enumerate}[leftmargin=*]
    \item Raw IPFIX features, such as IP protocol identifier, number and size of packets in a flow, destination network, flow end reason, etc.
    \item Features calculated using the raw features of the same IPFIX, such as flow IAT.
    \item DNS-related features, such as the percentage of numerical characters in the requested domain, etc.
    \item Reputation intelligence features~\cite{webroot}, such as the reputation status of the destination IP address, which denotes whether the destination is likely to be associated with phishing attacks, spam, mobile threats, etc.
    \item `Time-based' features (which partially resemble the stateful features used in~\cite{Doshi2018MachineDevices}), calculated using the raw features of the preceding IPFIXs of the \emph{other} devices of the same IoT model. These two `time-based' features count the number of times during the previous complete hour in which any of the \emph{other} devices contacted \emph{(1)} the network of the \emph{current} IPFIX's destination IP address and \emph{(2)} its destination port, as another means of collaboration for anomaly/attack detection. The rationale is that an outbound flow whose destination network and/or port were not recently contacted by the counterparts of the inspected device could indicate an anomalous or even malicious behavior, e.g., communication with the C\&C (command and control) server of a botnet~\cite{hallman2017ioddos}.
\end{enumerate}

\subsection{Implemented attacks and corresponding labels}\label{subsec:implemented_attacks_and_Corresponding_Labels}

In our lab, we infected streamer.Amazon.Fire\_TV\_Gen\_3 with a cryptominer~\cite{cryptominer} and executed cryptomining from this device. To imitate a scanning activity typically performed by some botnets, we also scanned the network using Nmap (Network Mapper), as in~\cite{bou2014fingerprinting}. In accordance, we labeled these malicious activities as \emph{(1)} `is executing cryptomining,' or \emph{(2)} `being scanned by Nmap.' All of the remaining IPFIXs captured in our lab or on the home networks were labeled as `assumed benign,' mainly because we could not clearly substantiate the ground truth.

\subsection{Data merging, enrichment, preprocessing, and partitioning}\label{subsec:Data Enrichment_preprocessing_and_Partitioning}

The first stage of turning the raw data into a dataset ready for experimentation was selecting the \emph{outbound} IPFIXs in which the source was the monitored streamer.Amazon.Fire\_TV\_Gen\_3 devices (as opposed to inbound IPFIXs where these devices were the destination) and combining the selected outbound IPFIXs with their DNS requests and responses and their destinations' reputation intelligence data. Then we enriched the merged data by extracting features such as flow duration and IAT, DNS-related features, and (collaborative) time-based features.  Afterwards, as part of preliminary data preprocessing, we removed IPFIXs whose source MAC address or IAT were missing, removed the first two recorded hours of traffic data (because its time-based features could not be calculated properly), and in cases in which the time-based features were missing values, we input a value of zero.

After constructing the preliminary dataset and performing preprocessing, we labeled and partitioned it chronologically into three mutually-exclusive subsets, namely $F^m_{training}$, $F^m_{validation}$, and $F^m_{test}$ (13, 3, and 5 days, respectively, 21 days altogether). The total number of IPFIXs for each label in these data partitions is provided in Table~\ref{tab:daily_labels}. Then, as a final stage of preparing the dataset for experimentation, we performed several data selection and cleansing operations, mostly \emph{(1)} ensuring that $F^m_{test}$ only includes instances captured in our controlled lab (and thus could reliably be used to evaluate anomaly/attack detection performance), while \emph{(2)} ensuring that $F^m_{training}$ and $F^m_{validation}$ only include instances that were captured in \emph{another} network where streamer.Amazon.Fire\_TV\_Gen\_3 was deployed. This data selection strategy enabled us to evaluate the generalizability of our collaborative method and answer the question: Can an anomaly detector, which was trained and optimized using shared traffic data from multiple identical IoT devices deployed in various \emph{other} networks, effectively detect IoT-related cyber-attacks that involve an identical IoT device that was not represented in $F^m_{training}$ or $F^m_{validation}$? Additional data cleansing operations included \emph{(3)} keeping only external communications (i.e., communication sent outside each device's network)~\cite{Pashamokhtari2021Inferring}; \emph{(4)} keeping only assumed-benign flows in $F^m_{training}$ and $F^m_{validation}$, as opposed to $F^m_{test}$ which contains both malicious and assumed-benign flows; and \emph{(5)} sanitizing $F^m_{training}$~\cite{cretu2008casting,8418594} in order to avoid model poisoning~\cite{rey2022federated} by removing extreme outliers, i.e., excluding training flows whose destination ports are too scarce, collaboratively considering all of the IoT devices in $F^m_{trn}$. 

\begin{figure}[!h]
\centering
\includegraphics[width=0.5\textwidth, trim={0 0cm 0 0cm},clip]{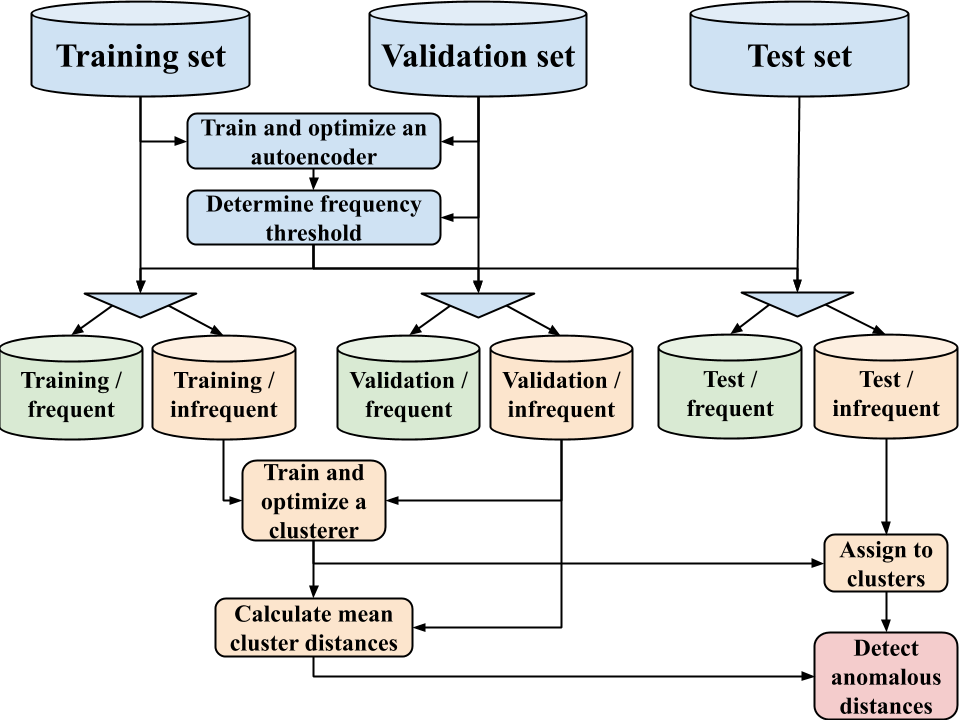}
\caption{Data partitioning and analytical flow for evaluation of our method}
\label{fig:evaluation_partitioning}
\end{figure}

The entire process of data partitioning and analytical flow is illustrated in Fig.~\ref{fig:evaluation_partitioning}. To remove ambiguity and enable research reproducibility, we share the final cleansed and preprocessed dataset with the public, anonymized and ready for experimentation.

\begin{table*}[ht]
\centering
\caption{Metrics used to evaluate our method}
\label{tab:metrics_used}
\resizebox{\textwidth}{!}{
\bgroup
\def\arraystretch{1.3}
\begin{tabular}{l|c|l|c}
\hhline{====}
\multicolumn{1}{c|}{\textbf{Metric}}                  & \multicolumn{1}{c|}{\textbf{Abbreviation}}         & \multicolumn{1}{c|}{\textbf{Calculation}}                               & \multicolumn{1}{c}{\textbf{Coverage}}            \\ \hhline{====}
True positives                                 & TP                                        & $\#(actual = malicious,\:predicted = `malicious')$               & \multirow{9}{*}{\begin{tabular}[t]{@{}l@{}}A given\\attack\\scenario\end{tabular}} \\ \cline{1-3}
False positives                                & FP                                        & $\#(actual = benign,\:predicted = `malicious')$                  &                                          \\ \cline{1-3}
True negatives                                 & TN                                        & $\#(actual = benign,\:predicted = `benign')$                     &                                          \\ \cline{1-3}
False negatives                                & FN                                        & $\#(actual = malicious,\:predicted = `benign')$                  &                                          \\ \hhline{===}
False positive rate                          & FPR                                       & $\frac{FP}{(FP+TN)}$                                                   &                                          \\ \cline{1-3}
Precision                                     &                                           & $\frac{TP}{(TP+FP)}$                                                   &                                          \\ \cline{1-3}
Recall                                        &                                           & $\frac{TP}{(TP+FN)}$                                                   &                                          \\ \cline{1-3}
F1 score                                      & F1                                        & $2 \cdot \frac{precision \cdot recall}{precision + recall}$                    &                                          \\ \cline{1-3}
Area under the precision-recall curve         & AUPRC                                     &                                                                &                                          \\ \hhline{====}
Macro-average \textless{}metric\textgreater{} & & $\frac{\sum <metric>}{\#(attack\:scenarios)}$ & Multiple attack scenarios                     \\ \hhline{====}
\end{tabular}
\egroup
}
\end{table*}

\subsection{Hyperparameter tuning}\label{subsec:hyp_param_tuning}

In our experiments we sought to optimize the following five hyperparameters that we suspected would have an effect on our method's performance, as follows:
\begin{enumerate}[leftmargin=*]
    \item Destination IP addresses have been used extensively in past research~\cite{guo2018ip} for IoT device fingerprinting, since IoT devices regularly communicate with a limited set of (their manufacturers') servers. However, when one-hot encoding the destination IP address, dimensionality and sparsity are likely to increase and thus harm model performance. Moreover, while a domain name typically remains unchanged, the associated IP addresses can change, so the model generalizability might decrease over time. To address these challenges, in our experiments we tested whether completely disregarding the IP addresses improves performance. We also tested the effect of using (and then one-hot encoding) only the prefixes of the destination IP addresses, which indicate the networks of the hosts~\cite{liu2020tpii,meidan2020deNAT}.
    \item Different transformations of numerical features were previously found~\cite{jiang2008transformations} to have varying levels of impact on model performance, depending on the dataset and algorithm selected. In our experiments we chose to compare the use of numerical features as is vs. log-transforming them.
    \item As noted in Subsection~\ref{subsec:filter_1_frequent}, we expect most of the normal outbound flows of an IoT device (presumably around $\frac{2}{3}$) to be relatively stable. So, in order to set $Filter^m_1$'s threshold $TH^m_{frequent}$ we compared between choosing the $60^{th}$ or the $70^{th}$ percentile of the MSE using $F^m_{validation}$. Lower thresholds are expected to increase the number of FPs using $F^m_{test}$, while higher values are likely to increase the number of FNs.
    \item Having accumulated a large variety of features in the dataset, we decided to compare four options of the feature set to be used to implement the cluster-based $Filter_2^m$: \emph{(1)} all of the available features, \emph{(2)} a manually selected feature subset which is comprised of the `immediate suspects,' namely octet\_delta\_count, avg\_packet\_size, flow\_duration\_milliseconds, same\_dest\_ip\_count\_pool, and same\_dest\_port\_count\_pool (see Appendix~\ref{appendix:features} for feature descriptions), \emph{(3)} a PCA transformation of the original features, and \emph{(4)} the hidden layer of the AE underlying $Filter_1^m$, as in~\cite{HASEEB2022487} and~\cite{yeom2020autoencoder}.
    \item In $Filter_2^m$, the distance of a flow $f$ from the cluster to which it is assigned is used as a measure of abnormality. In our experiments we compared two approaches for calculating this distance: \emph{(1)} the raw Euclidean distance between $f$ and the cluster centroid, and \emph{(2)} the normalized Euclidean distance, i.e., the number of standard deviations from the mean. In order to restrict these distance measures to the range [0, 1], as in probability, we eventually compared $tanh(raw\:Euclidean\:distance)$ with $tanh(normalized\:Euclidean\:distance)$.
\end{enumerate}

\section{Results}\label{sec:results}

In this section, we present the results of the hyperparameter tuning process, providing the hyperparameter values that resulted in the best performance using $F^m_{test}$. Then, using the best-performing combination of hyperparameter values, we present the results obtained by $Filter^m_1$ and $Filter^m_2$ (Subsections~\ref{subsec:results_for_filter_1} and~\ref{subsec:results_for_filter_2}, respectively) using $F^m_{test}$. In Subsection~\ref{subsec:overall_results}, we summarize the overall empirical results of our two-step method and compare these results to those of classical one-step anomaly detection algorithms, and in Subsection~\ref{subsec:sensitivity_analysis}, we analyze the sensitivity of our method to a high-importance training consideration, namely the quantity of most recent IPFIXs present in the training set. Throughout this section we evaluate our results using the performance metrics defined in Table~\ref{tab:metrics_used}.

\subsection{Hyperparameter tuning results}\label{subsec:results_hyp_param_tuning}

Table~\ref{tab:hyp_params} lists the five hyperparameters tuned throughout our experiments (described in Subsection~\ref{subsec:hyp_param_tuning}) and indicates the value that performed best using $F^m_{test}$ in terms of the macro-average AUPRC (area under the precision-recall curve). We denote this best-performing combination as $hyp^*$. We chose to use the macro-average, because our dataset is imbalanced with respect to the malicious labels. We reached $hyp^*$ after conducting an exhaustive grid search, where the duration of a single end-to-end experiment was approximately $48\pm24$ minutes ($Mean \pm St.Dev.$), using the experimental setup described in Subsection~\ref{subsec:experimental_setup}.

\begin{table}[!h]
\centering
\caption{Hyperparameter tuning performed in our experiments}
\label{tab:hyp_params}
\resizebox{\columnwidth}{!}{
\bgroup
\def\arraystretch{1.4}
\begin{tabular}{l|l|c}
\hhline{===}
\multicolumn{1}{c|}{\textbf{Hyperparameter}}                                      & \multicolumn{1}{c|}{\textbf{Evaluated values}} & \textbf{\begin{tabular}{c}Performed\\best\end{tabular}} \\ \hhline{===}
\multirow{2}{*}{\begin{tabular}[t]{@{}l@{}}Treatment of the\\destination IP address\end{tabular}}                            & Remove completely                               & \Checkmark                       \\ \cline{2-3} 
                                                                                    & Use only the prefix (network)                       &                         \\ \hline
\multirow{2}{*}{Treatment of numeric features}                                      & Log transform                                   & \Checkmark                       \\ \cline{2-3} 
                                                                                    & Use as is                                       &                         \\  \hline
\multirow{2}{*}{\begin{tabular}[t]{@{}l@{}}MSE threshold between `frequent'\\and `infrequent' ($Filter_1^m$)\end{tabular}}         & $60^{th}$ percentile using $F^m_{validation}$     & \Checkmark                       \\ \cline{2-3} 
                                                                                    & $70^{th}$ percentile using $F^m_{validation}$     &                         \\ \hline
\multirow{4}{*}{\begin{tabular}[t]{@{}l@{}}Features used for clustering\\($Filter_2^m$)\end{tabular}}                                & All available features                          & \Checkmark                       \\ \cline{2-3} 
                                                                                    & AE's hidden layer                               &                         \\ \cline{2-3} 
                                                                                    & PCA                                             &                         \\ \cline{2-3} 
                                                                                    & Manually selected feature subset                 &                         \\ \hline
\multirow{2}{*}{\begin{tabular}[t]{@{}l@{}}Cluster distance to differentiate between\\`known' and `unknown' ($Filter_2^m$)\end{tabular}} & Centroid distance                               & \Checkmark                       \\ \cline{2-3} 
                                                                                    & Normalized distance from the mean               &                         \\ \hhline{===}
\end{tabular}
\egroup
}
\end{table}

\begin{figure*}[!ht]
\begin{minipage}{.45\linewidth}
\subfloat[]{\label{subfig:hyp_dest_IP}\includegraphics[height=0.2\textheight, trim={0cm 0cm 0cm 0cm},clip]
{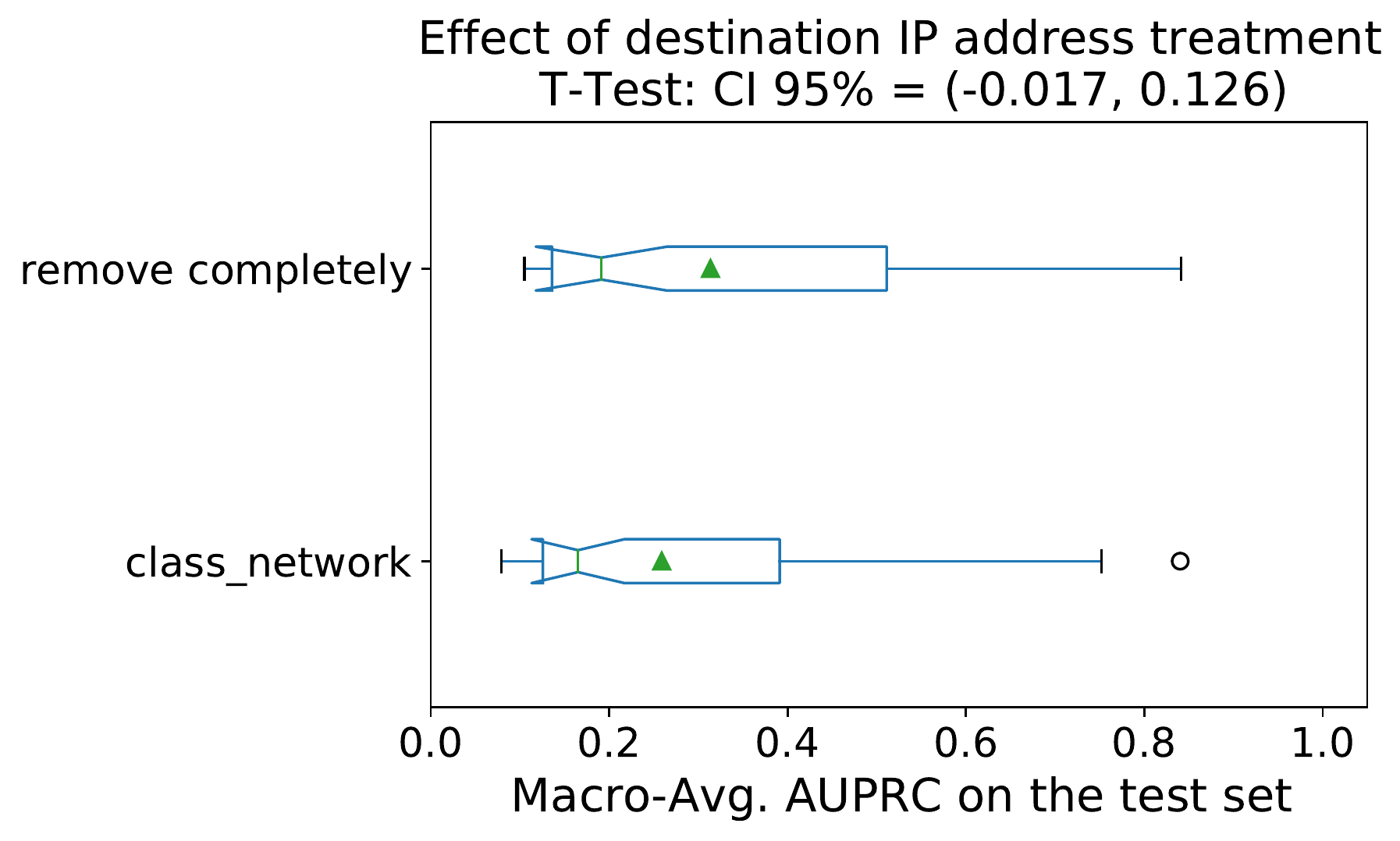}}
\hspace{\fill}
\end{minipage}
\hspace{\fill}
\begin{minipage}{.45\linewidth}
\centering
\subfloat[]{\label{subfig:hyp_numerics}\includegraphics[height=0.2\textheight, trim={0cm 0cm 0cm 0cm},clip]
{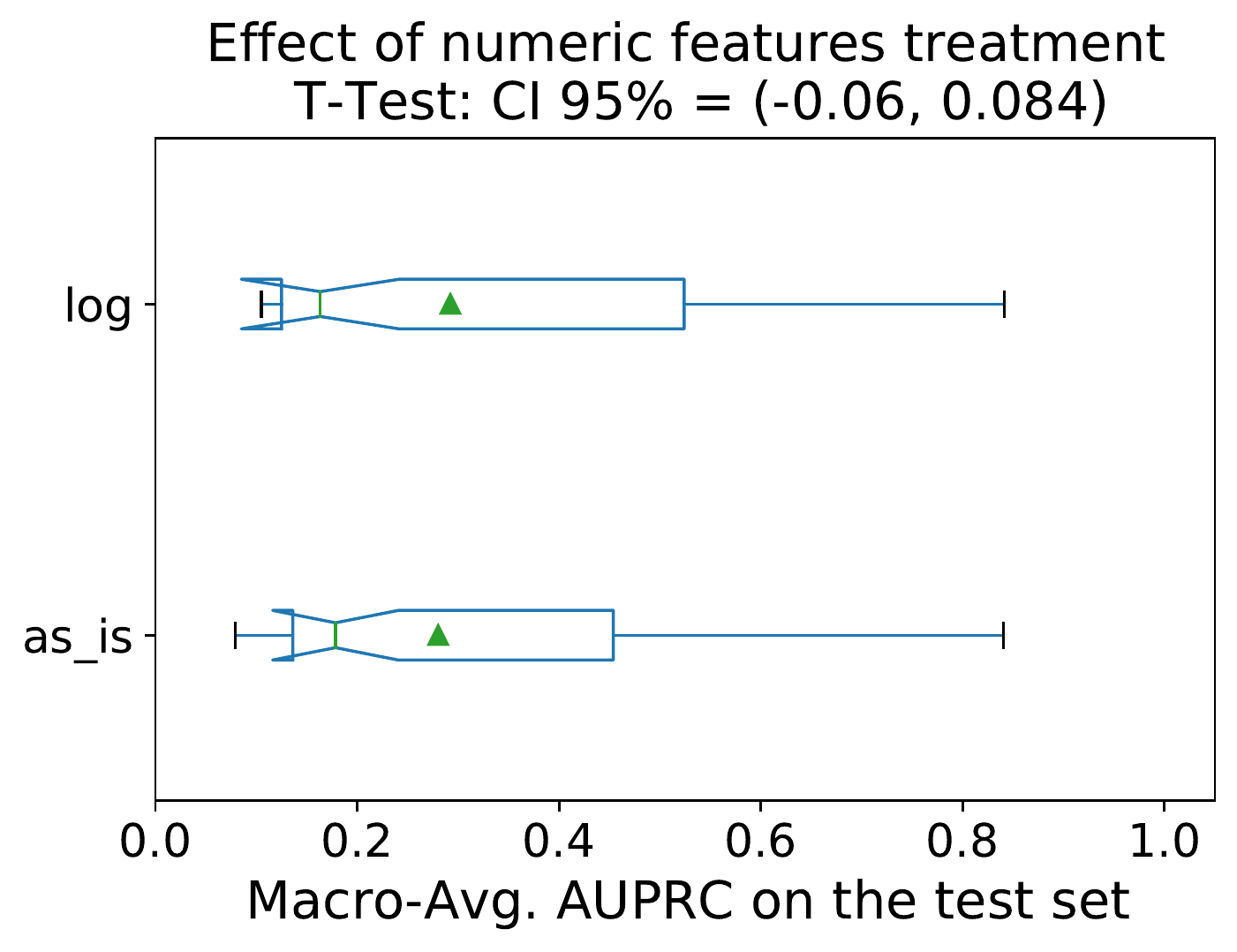}}
\end{minipage}
\hspace{\fill}
\vskip\baselineskip
\begin{minipage}{.45\linewidth}
\hspace{\fill}
\subfloat[]{\label{subfig:hyp_frequency_th}\includegraphics[height=0.2\textheight, trim={0cm 0cm 0cm 0cm},clip]
{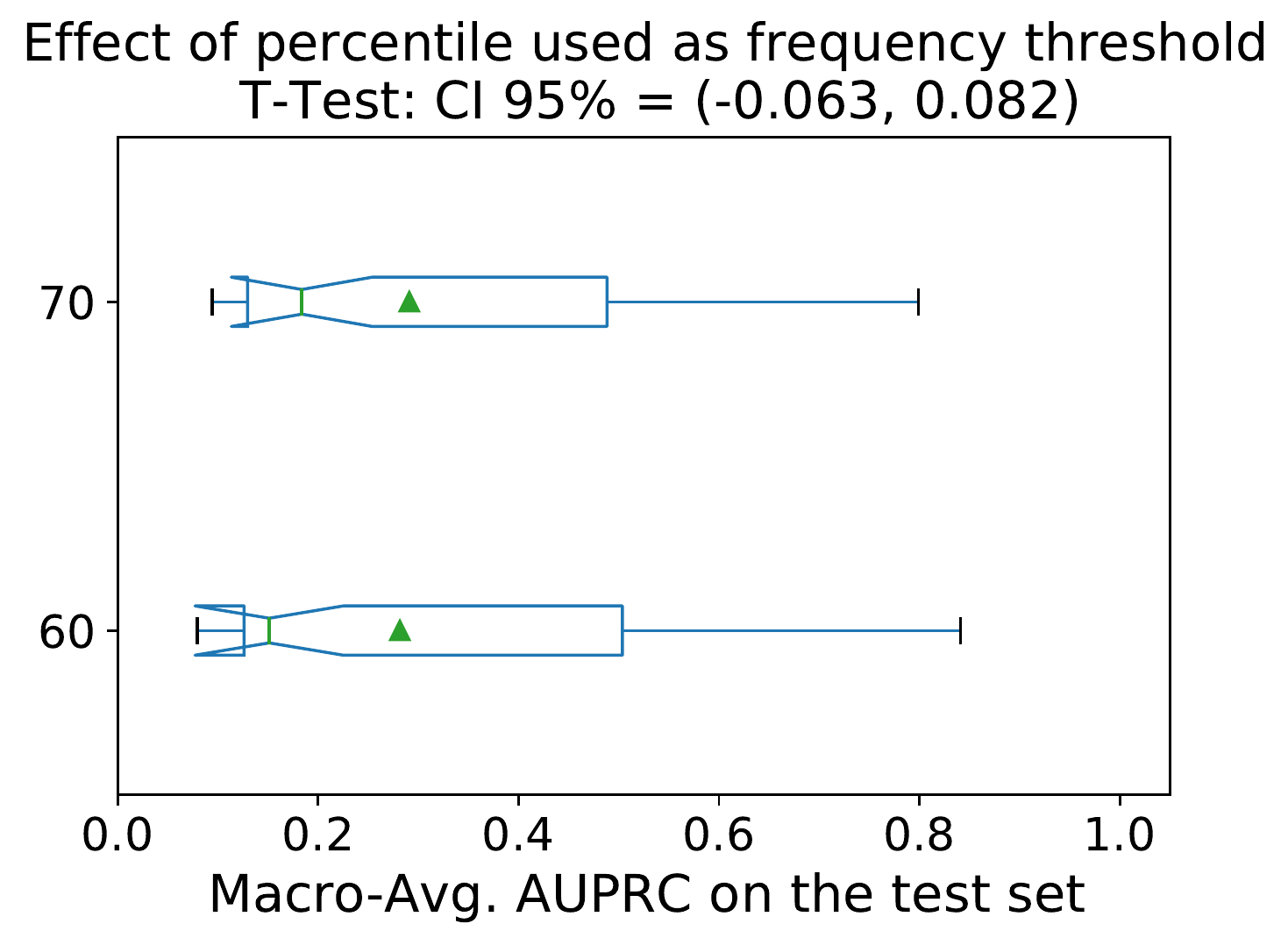}}
\end{minipage}
\begin{minipage}{.45\linewidth}
\centering
\subfloat[]{\label{subfig:hyp_clustering_features}\includegraphics[height=0.2\textheight, trim={0cm 0cm 0cm 0cm},clip]
{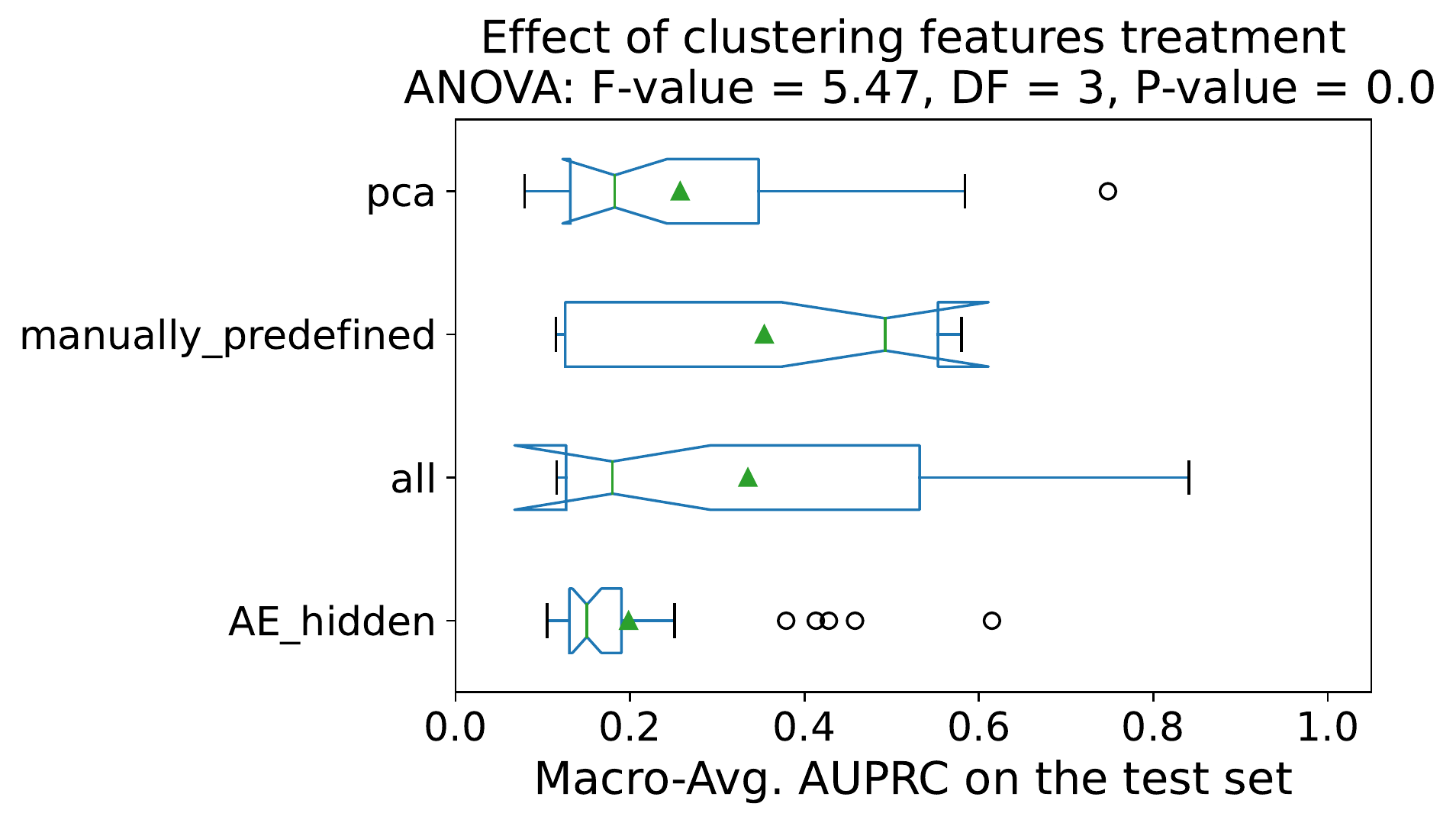}}
\end{minipage}
\hspace{\fill}
\vskip\baselineskip
\hspace{\fill}
\begin{minipage}{.45\linewidth}
\subfloat[]{\label{subfig:hyp_normality_method)}\includegraphics[height=0.2\textheight, trim={0cm 0cm 0cm 0cm},clip]
{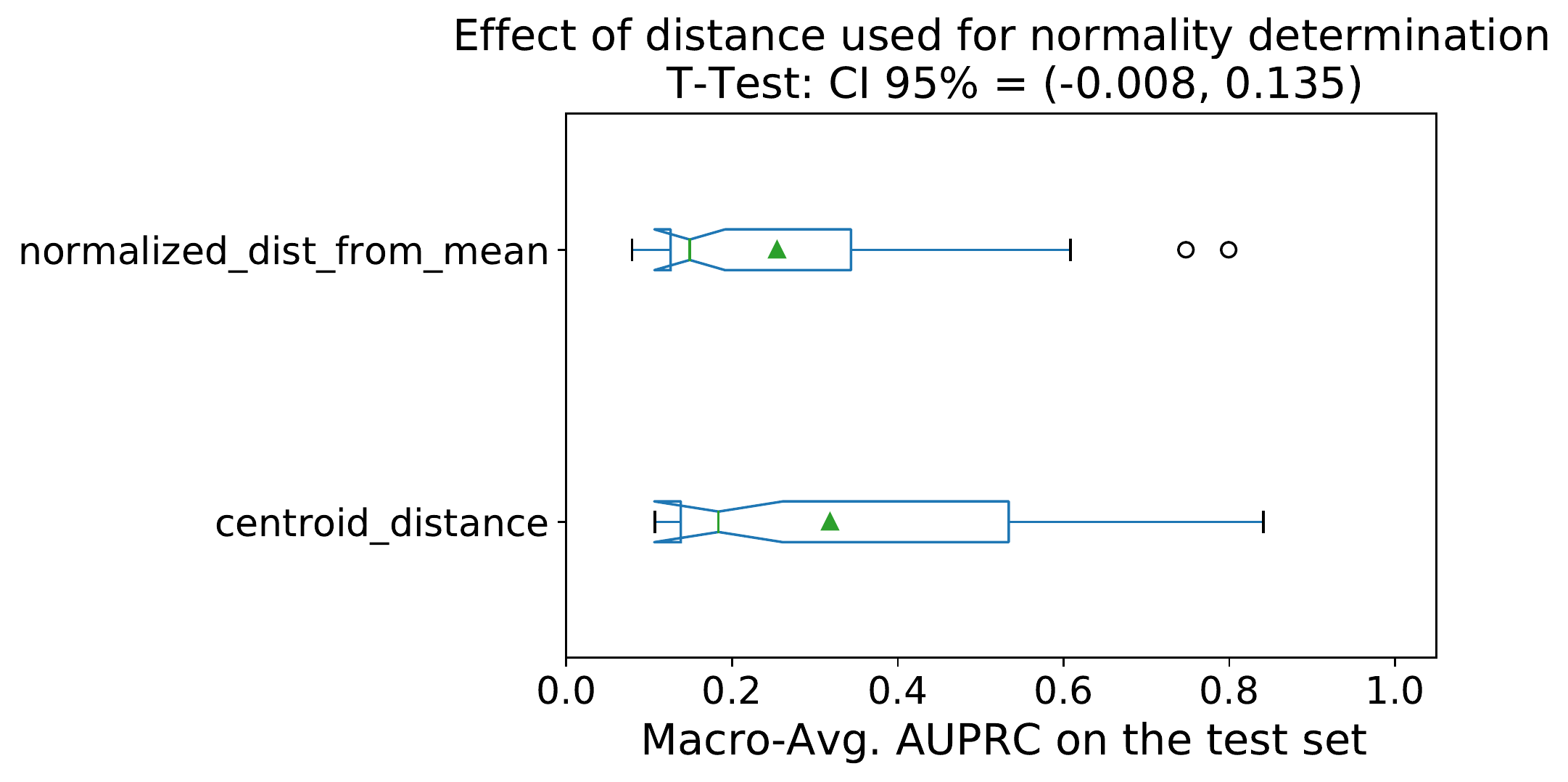}}
\hspace{\fill}
\end{minipage}
\hspace{\fill}
\caption{Effect of the various tested hyperparameters on the macro-average AUPRC using $F^m_{test}$, estimated through 95\% confidence intervals (CIs) and ANOVA}
\label{fig:hyp_params}
\end{figure*}

Based on the grid search we conducted, the macro-average AUPRC using $F^m_{test}$ was $0.286\pm0.206$, out of which the highest value (0.841) was attained by using the $hyp^*$ values presented in Table~\ref{tab:hyp_params}. Overall, as can be seen in Fig.~\ref{fig:hyp_params}, four of the five hyperparameters tuned had no significant effect on the  macro-average AUPRC using $F^m_{test}$. The fifth hyperparameter was the clustering feature treatment. An ANOVA test found a significant difference in means ($P-value=0.0$ (see Fig.~\ref{subfig:hyp_clustering_features})). This ANOVA test was followed by pairwise Tukey-HSD post-hoc tests which found that using the hidden AE layer is inferior to using all of the original features ($P-value=0.034$) and to the manually-predefined subset of the original features ($P-value=0.011$).

\begin{figure}[ht]
\centerline{\includegraphics[width=0.8\linewidth, trim={7cm 4cm 8cm 5.6cm},clip]{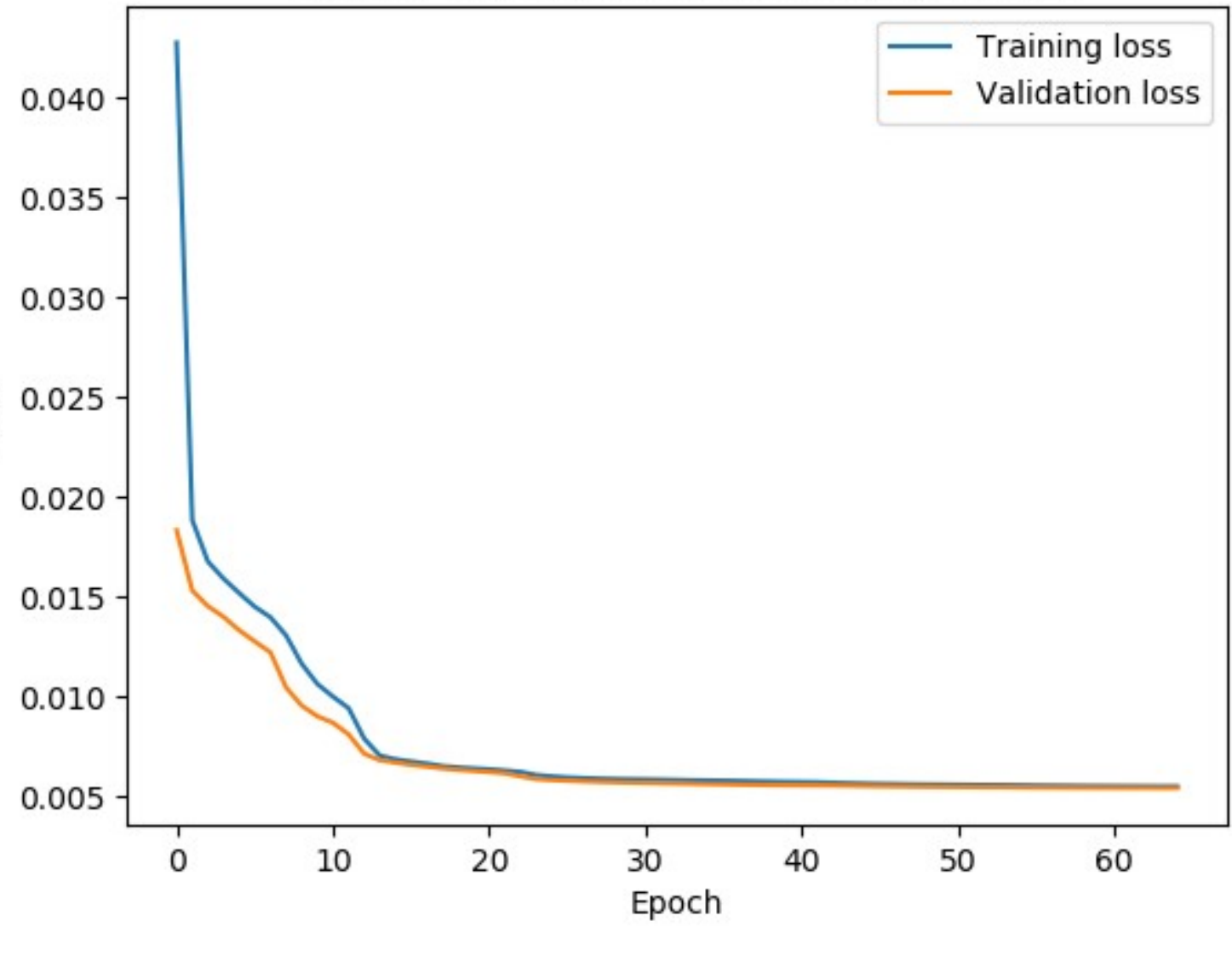}}
\caption{MSE (loss) convergence during training using $hyp^*$}
\label{fig:mse_trn_vld}
\end{figure}

\subsection{Results for $Filter^m_1$}\label{subsec:results_for_filter_1}

We used the Keras library~\cite{chollet2015keras} to implement $Filter_1^m$ as an AE with three hidden layers, i.e., a total of five dense fully-connected layers, including the input and output layers. We set the number of neurons in those layers to be 100-50-25-50-100\% of the input dimension. ReLU was used in all of the layers, except for the output layer, where we used a sigmoid activation function. We used a batch size of 64 and the Adam optimizer~\cite{Kingma2015AdamAM}. Although we allowed $epochs_{max}=200$, in most cases the training of $Filter_1^m$ stopped much earlier: $37.8 \pm 21.1$  epochs due to meeting the early stopping criteria: $delta_{min}=0.00001,\:patience_{max}=5$. The rapid convergence when using $hyp^*$ can be seen in Fig.~\ref{fig:mse_trn_vld}.

\begin{table}[!h]
\centering
\caption{MSE distribution across partitions}
\label{table:mse_partitions}
\resizebox{\columnwidth}{!}{
\bgroup
\def\arraystretch{1.3}
\begin{tabular}{l|c|c|c|c}
\hhline{=====}
\multicolumn{1}{c|}{\textbf{Percentile}} & \textbf{$MSE^{training}$} & \textbf{$MSE^{validation}$} & \textbf{$MSE^{test}_{benign}$} & \textbf{$MSE^{test}_{malicious}$} \\ \hhline{=====}
Min.                   & 0.0000                & 0.0000                  & 0.0000                   & 0.0010                      \\ \hline
25\%                   & 0.0002                & 0.0002                  & 0.0002                   & 0.0546                      \\ \hline
50\%                   & 0.0004                & 0.0003                  & 0.0004                   & 0.0565                      \\ \hline
75\%                   & 0.0011                & 0.0010                  & 0.0226                   & 0.0565                      \\ \hline
Max.                   & 0.1801                & 0.1802                  & 0.1960                   & 0.1984                      \\ \hhline{=====}
\end{tabular}
\egroup
}
\end{table}

The results presented in Table~\ref{table:mse_partitions} show that for the most part, the MSE of the trained AE-based $Filter^m_1$ is very low for benign IPFIXs, indicating the ability of $Filter^m_1$ to accurately reconstruct benign inputs. Moreover, the MSE on benign test records is almost identical to the MSE on benign training and validation records. This indicates the ability of $Filter^m_1$ to generalize to data on which it was not trained. It is also worth noting that the $25^{th}$ and $50^{th}$ percentiles of $MSE_{malicious}^{test}$'s distribution are much larger than the respective values on $MSE_{benign}^{test}$ by more than two orders of magnitude. This bolsters the very basic assumption behind anomaly-based NIDSs, namely that there is an association between (network traffic) data anomalies and malicious activity.

To differentiate between a frequent and infrequent IPFIX, we set the threshold $TH^m_{frequent}$ as the $60^{th}$ percentile of the MSE's distribution on $F^m_{validation}$. Using this threshold, as can be seen in Table~\ref{tab:filters_1_2}, $Filter^m_1$ did not produce any FNs on $F^m_{test}$, meaning that none of the actually malicious IPFIXs were mistakenly labeled by $Filter_1^m$ as frequent. The 10,489 IPFIXs labeled as infrequent, which could actually be `rare yet benign' or `malicious', proceeded to $Filter^m_2$ for further labeling as known or unknown, respectively. 

\begin{table}[!h]
\centering
\caption{Number of predicted vs. actual labels for $Filter_1^m$ and $Filter_2^m$ using $F^m_{test}$ and $hyp^*$}
\label{tab:filters_1_2}
\resizebox{\columnwidth}{!}{
\bgroup
\def\arraystretch{1.3}
\begin{tabular}{l||c|r|r||r|r|r}
\hhline{=======}
\multicolumn{1}{c||}{\multirow{3}{*}{\textbf{Actual}}} & \multicolumn{6}{c}{\textbf{Predicted}}                                                  \\ \cline{2-7} 
\multicolumn{1}{c||}{}                                 & \multicolumn{3}{c||}{\textbf{$Filter_1^m$}}      & \multicolumn{3}{c}{\textbf{$Filter_2^m$}} \\ \cline{2-7} 
\multicolumn{1}{c||}{} &
  \textbf{Frequent} &
  \multicolumn{1}{c|}{\textbf{Infrequent}} &
  \multicolumn{1}{c||}{\textbf{Total}} &
  \multicolumn{1}{c|}{\textbf{Known}} &
  \multicolumn{1}{c|}{\textbf{Unknown}} &
  \multicolumn{1}{c}{\textbf{Total}} \\ \hhline{=======}
Assumed benign                                        & \multicolumn{1}{r|}{11,983} & \textbf{10,489} & 22,472 & 10,174       & 315         & \textbf{10,489}      \\ \hline
Being scanned by Nmap                                 & \multicolumn{1}{r|}{-}      & 3,080  & 3,080  & 48           & 3,032       & 3,080       \\ \hline
Is executing cryptomining                             & \multicolumn{1}{r|}{-}      & 1,703  & 1,703  & -            & 1,703       & 1,703       \\ \hhline{=======}
\end{tabular}
\egroup
}
\end{table}

\subsection{Results for $Filter^m_2$}\label{subsec:results_for_filter_2}

\begin{figure}[ht]
\centerline{\includegraphics[height=0.2\textheight, trim={0cm 0cm 0cm 0cm},clip]
{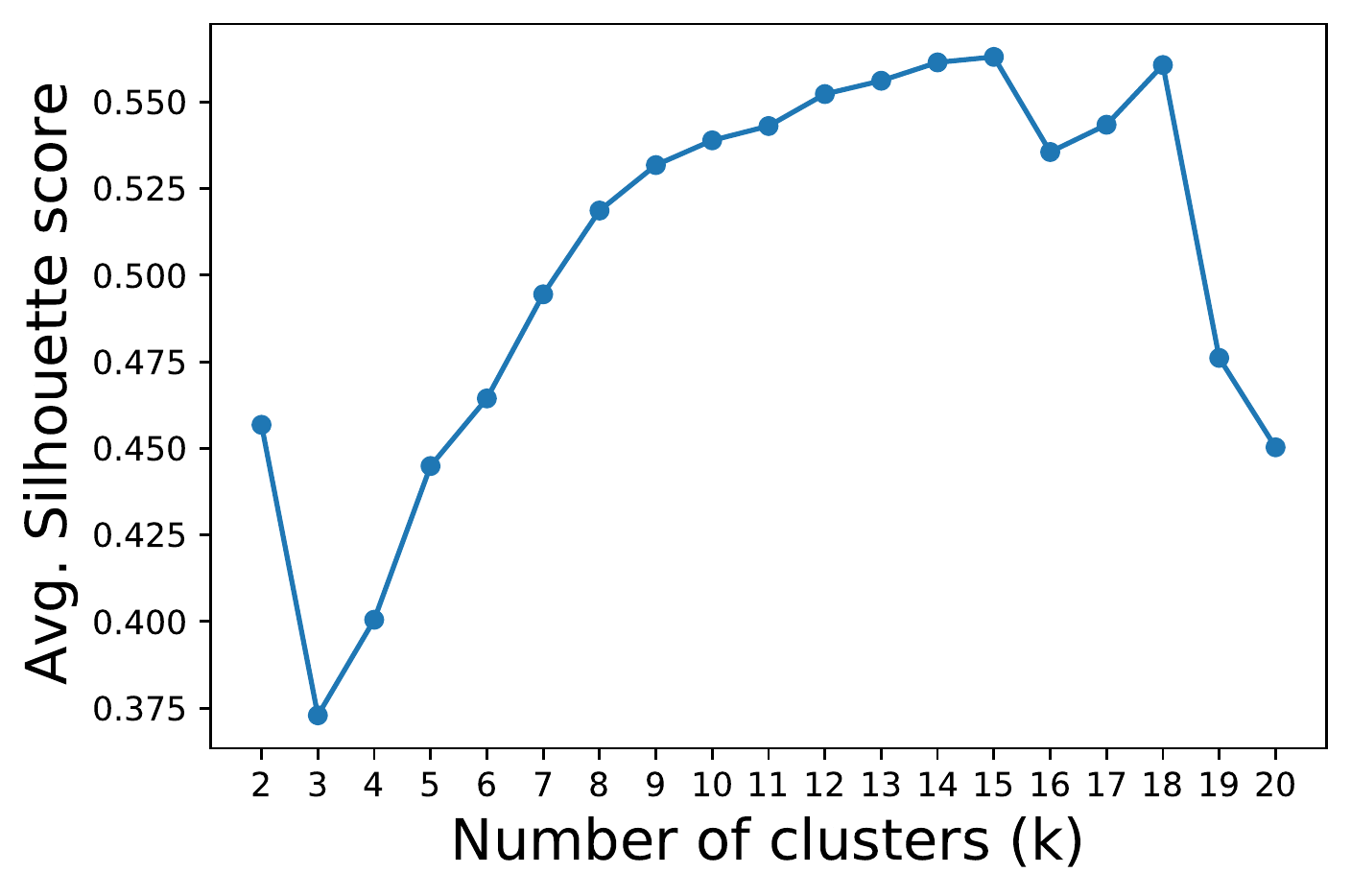}}
\caption{Determination of
\textit{k} based on the average silhouette score using $F^{m,\:infrq}_{validation}$}
\label{fig:subfig:silhouette_vs_k}
\end{figure}

\begin{figure}[h]
\centerline{\includegraphics[width=0.9\linewidth, trim={5.5cm 3.8cm 4cm 5.2cm},clip]
{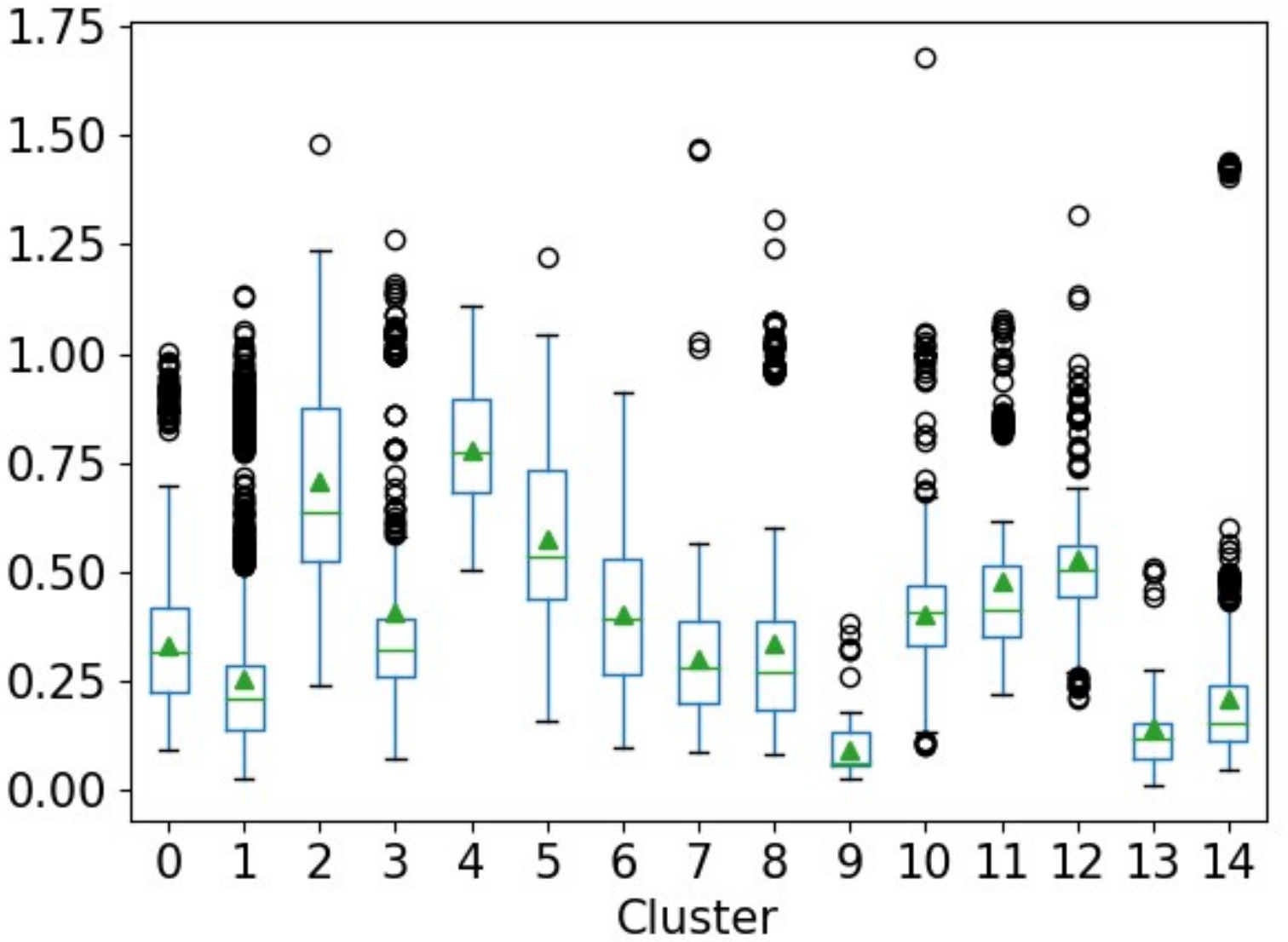}}
\caption{Distribution of the distance from each cluster in $F^{m,\:infrq}_{validation}$}
\label{subfig:cluster_distances}
\end{figure}

To implement $Filter_2^m$, we used $k$-means clustering and found that within the range of [$k_{min}=2,\:k_{max}=20$] the number of clusters that maximizes the average silhouette score~\cite{silhouette2004} using $F^{m,\:infrq}_{training}$ is $k^*=15$ (see Fig.~\ref{fig:subfig:silhouette_vs_k}). To differentiate between a known and unknown behavior, we utilized the distance of the inspected IPFIX from the nearest cluster. To align this metric with a probability expression, we considered $tanh(cluster-distance) \approx p(abnormality)$. The distribution of the raw distances from the $k^*=15$ cluster centers is illustrated in Fig.~\ref{subfig:cluster_distances} for each cluster separately (ranging up to $\sim1.7$ using $F^{m,\:infrq}_{validation}$).

\begin{figure}[h]
\centerline{\includegraphics[width=0.9\linewidth, trim={5cm 4cm 5cm 5.3cm},clip]
{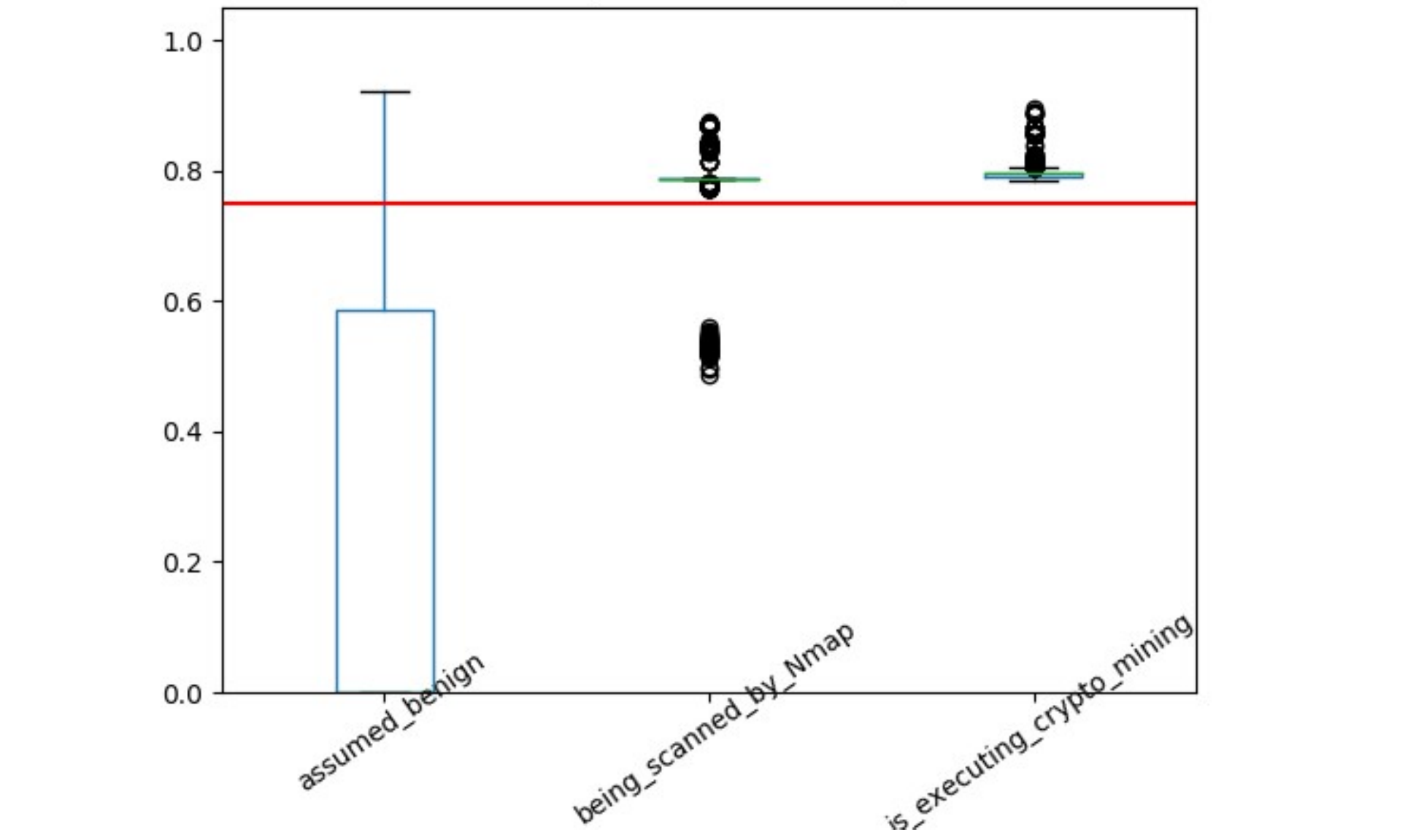}}
\caption{Tanh(distance) in $F^m_{test}$ to differentiate between known and unknown IPFIXs}
\label{subfig:tanh_cluster_distances}
\end{figure}

In contrast, as shown in Fig.~\ref{subfig:tanh_cluster_distances}, the distribution of $tanh(cluster-distance)$ using $F^m_{test}$ is bounded to within the range of $[0,\:1]$, where it can also be seen that a threshold of $0.75$ offers a good separation between most assumed-benign IPFIXs and malicious IPFIXs.

\begin{figure*}[ht]
\begin{minipage}{.49\linewidth}
\centering
\subfloat[]{\label{subfig:tanh_threshold_Nmap}\includegraphics[height=0.23\textheight, trim={0cm 0cm 0cm 0cm},clip]
{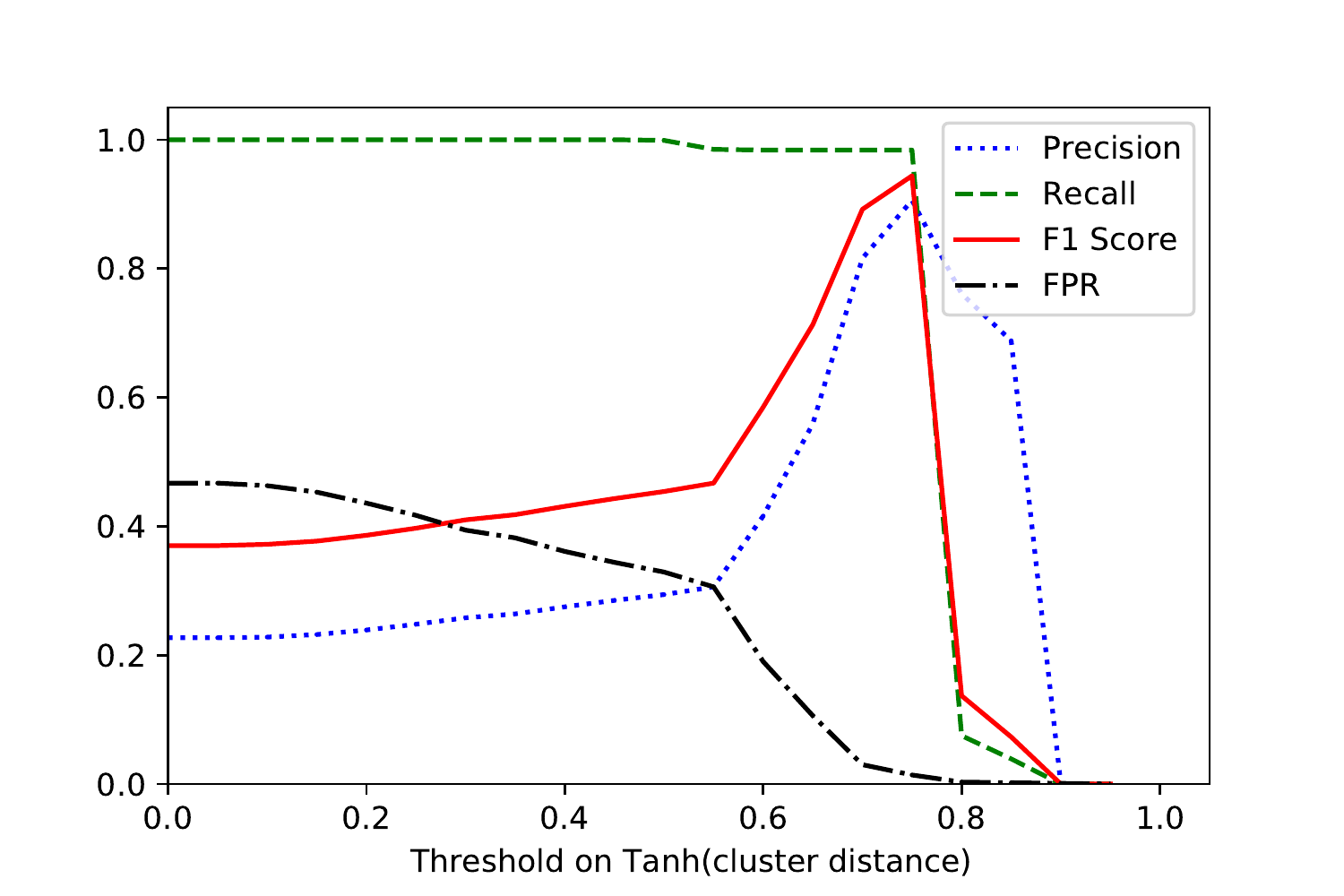}}
\end{minipage}
\begin{minipage}{.49\linewidth}
\centering
\subfloat[]{\label{subfig:tanh_threshold_crypto}\includegraphics[height=0.23\textheight, trim={0cm 0cm 0cm 0cm},clip]
{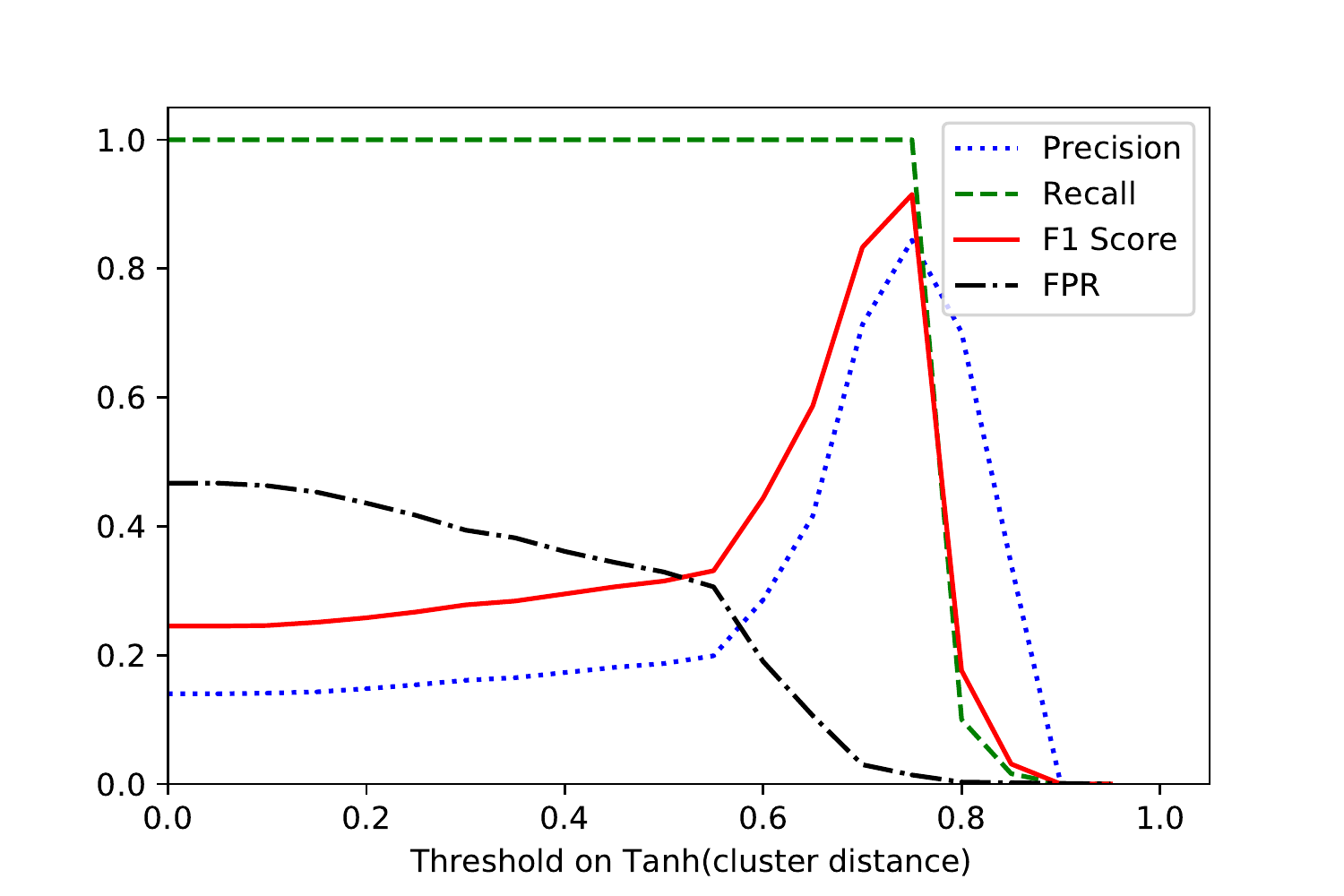}}
\end{minipage}
\caption{Performance of the KPIs as a function of the threshold on tanh(Euclidean distance from cluster centers) in $Filter_2^m$ for two attack scenarios: The IoT device is (a) being scanned by Nmap, and (b) executing cryptomining}
\label{fig:tanh_distance_threshold_selection}
\end{figure*}

\subsection{Overall results and benchmarking}\label{subsec:overall_results}

The KPIs which summarize the overall performance of our method using $F^m_{test}$ and $hyp^*$ are presented in Table~\ref{tab:macro_avg_KPIs}. The bottom row of the table shows that the sequential use of $Filter_1^m$ and $Filter_2^m$ achieves a 0.841 macro-average AUPRC. In contrast to this KPI, which considers a multitude of threshold values on $tanh(cluster-distance)$ in $Filter^m_2$, the top four rows relate only to one threshold, denoted as $TH^m_{known}$. As can be seen in Figs.~\ref{subfig:tanh_threshold_Nmap} and~\ref{subfig:tanh_threshold_crypto}, this threshold has a substantial impact on the precision, recall, F1 score, and FPR for the detection of Nmap scanning and cryptomining execution, respectively. Setting $TH^m_{known}=0.75$ yields a 0.929 macro-average F1 score. The recall is also very high (it is actually perfect for cryptomining execution detection), while the FPR has a relatively low value of 0.014 for both of the evaluated attacks.

\begin{table}[ht]
\centering
\caption{Summary of our method's KPIs using $F^m_{test}$ and $hyp^*$}
\label{tab:macro_avg_KPIs}
\resizebox{\columnwidth}{!}{
\bgroup
\def\arraystretch{1.3}
\begin{tabular}{l|c|c|c|c}
\hhline{=====}
\multicolumn{1}{c|}{\textbf{Metric}} & \textbf{\begin{tabular}{c}Being scanned\\by Nmap\end{tabular}} & \textbf{\begin{tabular}{c}Is executing\\cryptomining\end{tabular}} & \textbf{\begin{tabular}{c}Macro-\\average\end{tabular}} & \textbf{$TH^m_{known}$} \\ \hhline{=====}
Precision & 0.906 & 0.844 & 0.875 & \multirow{4}{*}{0.75}                      \\ \cline{1-4}
Recall    & 0.984 & 1.000 & 0.992 &                       \\ \cline{1-4}
F1 score        & 0.944 & 0.915 & 0.929 &                       \\
\cline{1-4}
FPR       & 0.014 & 0.014 & 0.014 &  \\
\hhline{=====}
AUPRC     & 0.827 & 0.855 & 0.841      & \multicolumn{1}{l}{}  \\ \hhline{====~}
\end{tabular}
\egroup
}
\end{table}

For benchmarking, we compared our two-step method with a variety of classical one-step algorithms widely used for anomaly and novelty detection in general~\cite{yang2020ddos,9166366,agarwal2021performance,villa2021semi,toshniwal2020overview}, and for IoT attack detection in particular~\cite{meidan2018nbaiot,qi2021detecting}. Two of these algorithms are the building blocks of CADeSH (namely, AE and $k$-means clustering, which we used to implement $Filter^m_1$ and $Filter^m_2$, respectively). The remaining three are OCSVM~\cite{scholkopf2001estimating}, LOF~\cite{breunig2000lof}, and IF~\cite{liu2008isolation}. This algorithm comparison (summarized in Table~\ref{tab:AUPRC_and_complexity_cadesh_vs_one_step}) takes into account both attack detection performance (in terms of AUPRC) and the upper bound of computational complexity (denoted using the \emph{big O notation}). As can be seen in Table~\ref{tab:AUPRC_and_complexity_cadesh_vs_one_step}, on macro-average, our two-step CADeSH method outperformed all of the one-step algorithms in terms of the AUPRC. That is, although $k$-means and OCSVM performed better than CADeSH in the detection of Nmap scanning, they performed much worse in detecting cryptomining execution, and vice versa; the AE performed better than CADeSH in the detection of cryptomining execution, however it performed (much) worse in detecting Nmap scanning. In terms of computational complexity, we note that CADeSH is relatively expensive. Still, it is our opinion that attack detection performance is much more crucial in real-life scenarios. Moreover, when solutions like parallel execution using, e.g., graphics processing units (GPUs) are widely available, the actual time differences between the algorithms become less and less significant.

\begin{table*}[ht]
\centering
\caption{Comparison of our two-step method's with various one-step novelty detection algorithms}
\label{tab:AUPRC_and_complexity_cadesh_vs_one_step}
\resizebox{\textwidth}{!}{
\bgroup
\def\arraystretch{1.3}
\begin{tabular}{l||ccc||ll}
\hhline{======}
\multicolumn{1}{c||}{}                & \multicolumn{3}{c||}{\textbf{Attack detection AUPRC using $F^m_{test}$ and $hyp^*$}}                                                                                   & \multicolumn{2}{c}{\textbf{Computational complexity \cite{7953382,8422402,RAZZAK2020715,bdcc5010001,4781136}}}                                                                                                                                                                                                 \\ \cline{2-6}
\multicolumn{1}{c||}{\textbf{Method}} & \multicolumn{1}{c|}{\textbf{Being scanned by Nmap}} & \multicolumn{1}{c|}{\textbf{Is executing cryptomining}} & \textbf{Macro-average} & \multicolumn{1}{c|}{\textbf{Upper bound}}                               & \multicolumn{1}{c}{\textbf{Terms}}                                                                                                                                                                                                                          \\ \hhline{======}
CADeSH                               & \multicolumn{1}{c|}{0.827}                          & \multicolumn{1}{c|}{0.855}                              & \textbf{0.841}         & \multicolumn{1}{l|}{$O(k\operatorname{-}Means)+O(AE)$}                                   & \multirow{6}{*}{\begin{tabular}[c]{@{}l@{}}$d$: number of dimensions (features)\\ $k$: number of clusters to train\\ $r$: number of records\\ $I$: number of iterations until convergence\\ $t$: number of trees in the ensemble\\ $s$: sub-sampling size\end{tabular}} \\ \cline{1-5}
k-Means                              & \multicolumn{1}{c|}{\textbf{0.970}}                 & \multicolumn{1}{c|}{0.167}                              & 0.569                  & \multicolumn{1}{l|}{$O(d \times k \times r \times I)$}                                            &                                                                                                                                                                                                                                                             \\ \cline{1-5}
AE                                   & \multicolumn{1}{c|}{0.176}                          & \multicolumn{1}{c|}{\textbf{0.917}}                     & 0.547                  & \multicolumn{1}{l|}{$O(d^2)$ for each layer}             &                                                                                                                                                                                                                                                             \\ \cline{1-5}
OCSVM                                & \multicolumn{1}{c|}{0.885}                          & \multicolumn{1}{c|}{0.129}                              & 0.507                  & \multicolumn{1}{l|}{$O(r \times d^2)$ for non-linear kernels} &                                                                                                                                                                                                                                                             \\ \cline{1-5}
LOF                                  & \multicolumn{1}{c|}{0.590}                          & \multicolumn{1}{c|}{0.240}                              & 0.415                  & \multicolumn{1}{l|}{$O(d^2)$}                            &                                                                                                                                                                                                                                                             \\ \cline{1-5}
IF                                   & \multicolumn{1}{c|}{0.338}                          & \multicolumn{1}{c|}{0.089}                              & 0.214                  & \multicolumn{1}{l|}{$O(t \times s \times log(s)$)}                                      &                                                                                                                                                                                                                                                             \\ \hhline{======}
\end{tabular}
\egroup
}
\end{table*}

\begin{figure}[!b]
\centerline{\includegraphics[width=1.0\linewidth, trim={0cm 0cm 0cm 0cm},clip]
{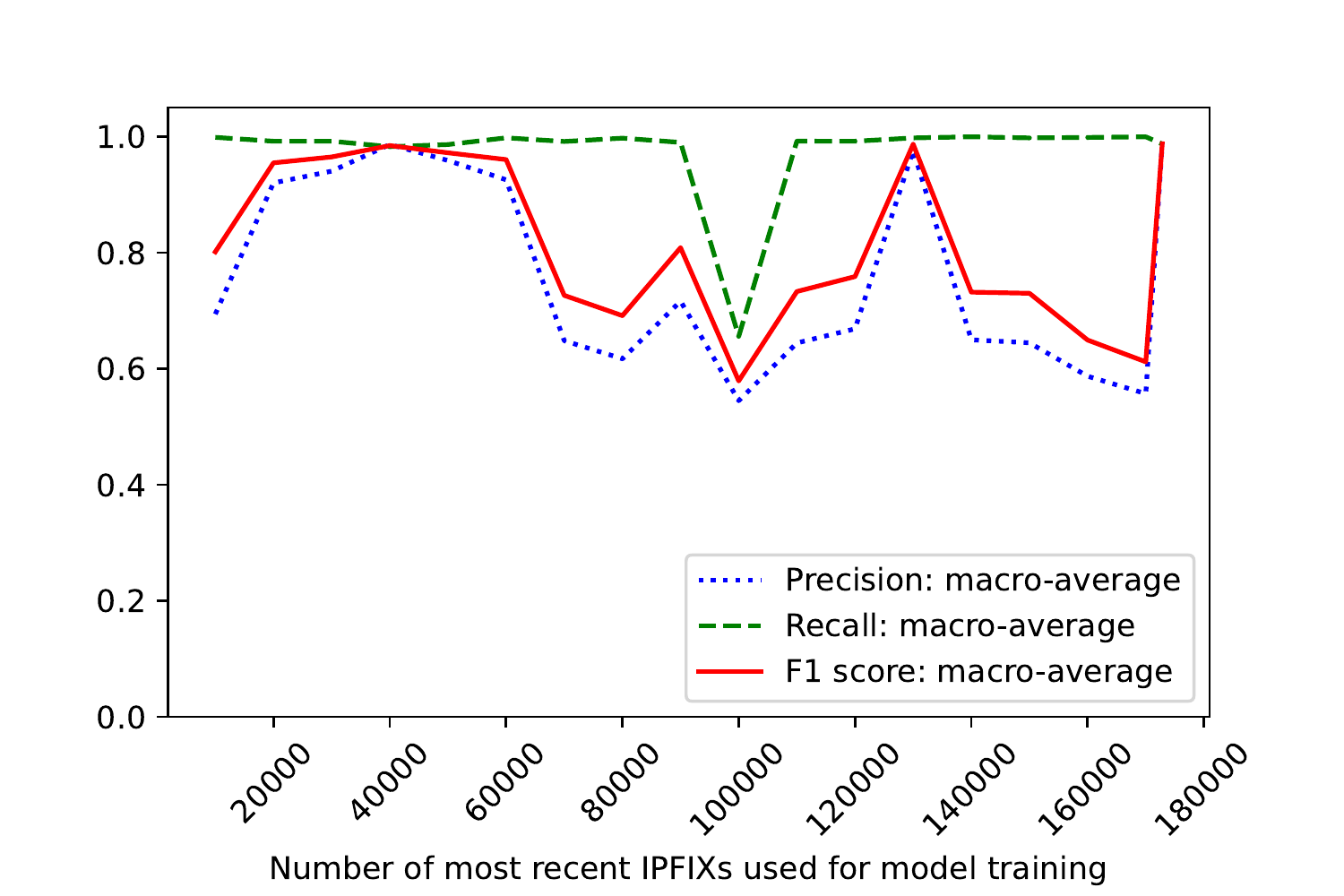}}
\caption{Sensitivity of CADeSH to the quantity of most recent training data}
\label{fig:sensitivity_analysis}
\end{figure}

\subsection{Sensitivity analysis}\label{subsec:sensitivity_analysis}

In our experiments, we also explored an issue that might need to be dealt with when considering real-world application of the proposed method, namely the quantity of training data required for (re-)training and how the value selected affects CADeSH's performance. The results of this experiment using $hyp^{*}$ and $F^m_{test}$ are presented in Fig.~\ref{fig:sensitivity_analysis}, where the macro-average precision, recall and F1 score are presented as a function of the number of most recent IPFIXs used for model training. In each iteration of this experiment, the selected IPFIXs were chronologically divided into a training and validation set, 80\% and 20\% of the data, respectively.

As can be seen in Fig.~\ref{fig:sensitivity_analysis}, although the best performance is attained using the entire training data (a total of 172,832 IPFIXs), an almost-identical performance level is attained using less than a quarter of this quantity (40,000 IPFIXs). On the one hand, smaller quantities of data might not represent enough human interactions and behaviors that are captured in the network traffic data, thus they lead to model under-fitting. On the other hand, in most cases larger quantities of training data lead to model over-fitting. A possible explanation is that these 40,000 IPFIXs (equivalent to approximately four days of traffic capturing) were the closest to the test set, such that they best represent the most recent normal activity based on which anomalies are detected. For example, if a given domain that is often contacted by model-$m$ IoT devices via DNS requests is not associated with a static IP address, then outdated data (from, e.g., over a week ago) might confuse the model.

\section{Discussion}\label{sec:discussion}

Despite its promising attack detection performance, our method has the following limitations. First, it requires the same data collection setup (hardware and/or software) across all of the monitored networks for efficient and reliable data sharing, and thus it might not be easy to scale. In our use case, the company we cooperated with supplied a piece of hardware to home subscribers as a third-party cyber-security service. This hardware enabled both the service of continuous smart home monitoring for malicious activities, and the standardized collection of data which can be shared and used for model training; however, it is possible that in the future ISPs would offer such a service themselves and preinstall the necessary software on the home routers they supply, instead of an additional hardware. Second, our method requires correct IoT model identification prior to anomaly detection, in order to enable the training and deployment of an anomaly detector for each IoT model separately; to this end, there are already promising research results regarding IoT identification based on IPFIX records, even behind a NAT~\cite{meidan2020deNAT,Pashamokhtari2021Inferring} (which is a typical setting in home networks). Third, as noted by~\cite{Pashamokhtari2021Inferring}, behavioral changes are not rare in the IoT, which could have an effect on the profile of normal network patterns. This might necessitate relatively frequent model retraining, depending on the degree to which the traffic patterns of each monitored IoT model $m$ are sensitive to behavioral changes. Yet, as described in Subsection~\ref{subsec:sensitivity_analysis}, only a small amount of (recent) training data is sufficient to produce a high-performing anomaly detector. Retraining might also be required following software updates, something which has yet to be empirically evaluated.

\section{Conclusion and future work}\label{sec:conclusion}

In this work, we proposed CADeSH, a novel collaborative two-step anomaly detection method. Using this method, an ISP can detect anomalous behaviors, which are indicative of malicious activities on specific IoT devices, locally (e.g., in a smart home). We evaluated our method empirically using real-world data from seven distinct streamer.Amazon.Fire\_TV\_Gen\_3 devices deployed in the networks of five real homes and interacted with naturally by different people over a period of 21 days. In our experiments we trained the anomaly detection models using data from those home networks (for which we did not have ground truth labels) and then tested the models using an identical device deployed in our controlled lab and carefully labeled this device's outbound traffic as `assumed benign,' `being scanned by Nmap' (imitating a botnet scan), or `is executing cryptomining.' These two well-known IoT-related cyber-attacks were detected by our method with a recall value of 0.929 (on macro-average), precision of 0.875, and an FPR as low as 0.014. A grid search conducted for hyperparameter tuning, followed by sensitivity analysis, revealed that training data of just 40,000 IPFIXs was enough to obtain excellent performance in this study.

Although in this work we focused on detecting anomalous behaviors locally, in the future our method could also be leveraged by an ISP to detect abnormal \emph{nationwide} trends. In other words, an IPFIX that is labeled as highly abnormal (i.e., representing a behavior which is both infrequent and unknown) is a strong indication of \emph{local} malicious activity. In comparison, a multitude of such abnormalities identified simultaneously on many IoT devices from a sufficiently large number of networks is likely to indicate the \emph{propagation} of a cyber-attack, e.g., botnet scanning or DDoS execution. We plan to design and evaluate this in our future work. Further extensions to this research include the evaluation of additional IoT device models, additional IoT-related attack scenarios, and additional algorithms and configurations for implementing $Filter^m_1$ and $Filter^m_2$, i.e., more complicated than a five-layer AE and $k$-means clustering.

\section*{Acknowledgment}

The authors would like to thank Dominik Breitenbacher, Tar Wolfson, Itamar Biton, Julia Tulisov, and Liam Habani for their valuable contributions to this research project.

\bibliographystyle{IEEEtran}
\bibliography{bb.bib}



\begin{IEEEbiography}[{\includegraphics[width=1in,height=1.25in,clip,keepaspectratio]{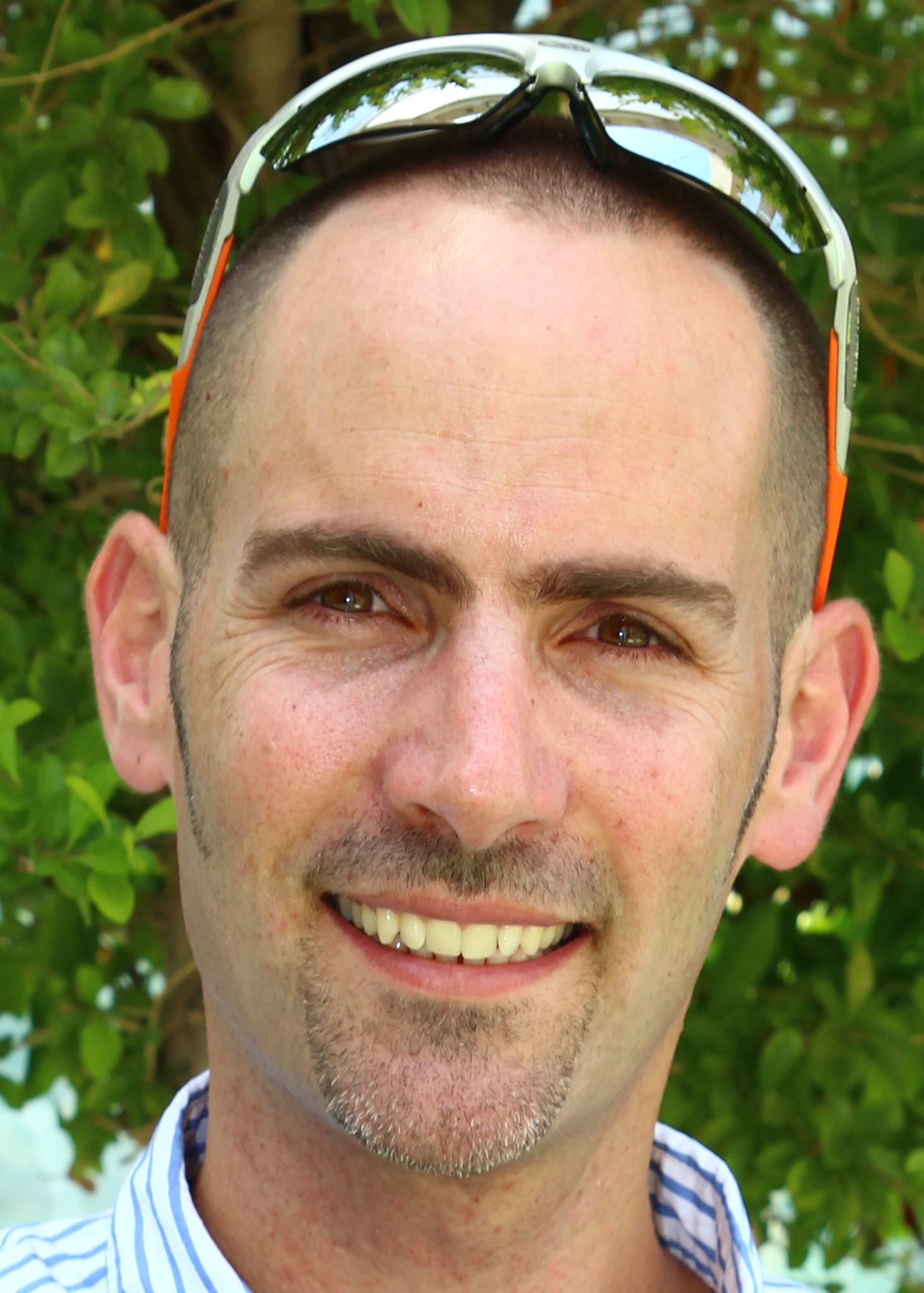}}]{Yair Meidan} is a Ph.D. candidate in the Department of Software and Information Systems Engineering (SISE) at Ben-Gurion University of the Negev (BGU). His research interests include machine learning and IoT security. Contact him at yairme@post.bgu.ac.il.
\end{IEEEbiography}

\begin{IEEEbiography}[{\includegraphics[width=1in,height=1.25in,clip,keepaspectratio]{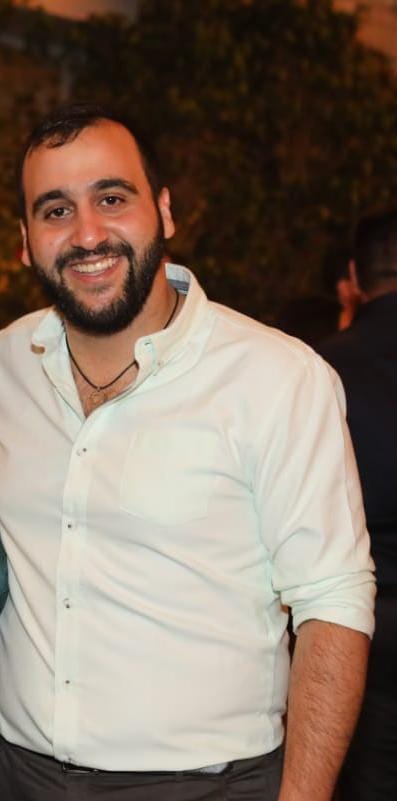}}]{Dan Avraham} is an M.Sc. student in the SISE Department at BGU. His research interests include machine learning and cyber-security. Contact him at danavra@post.bgu.ac.il.
\end{IEEEbiography}

\begin{IEEEbiography}[{\includegraphics[width=1in,height=1.25in,clip,keepaspectratio]{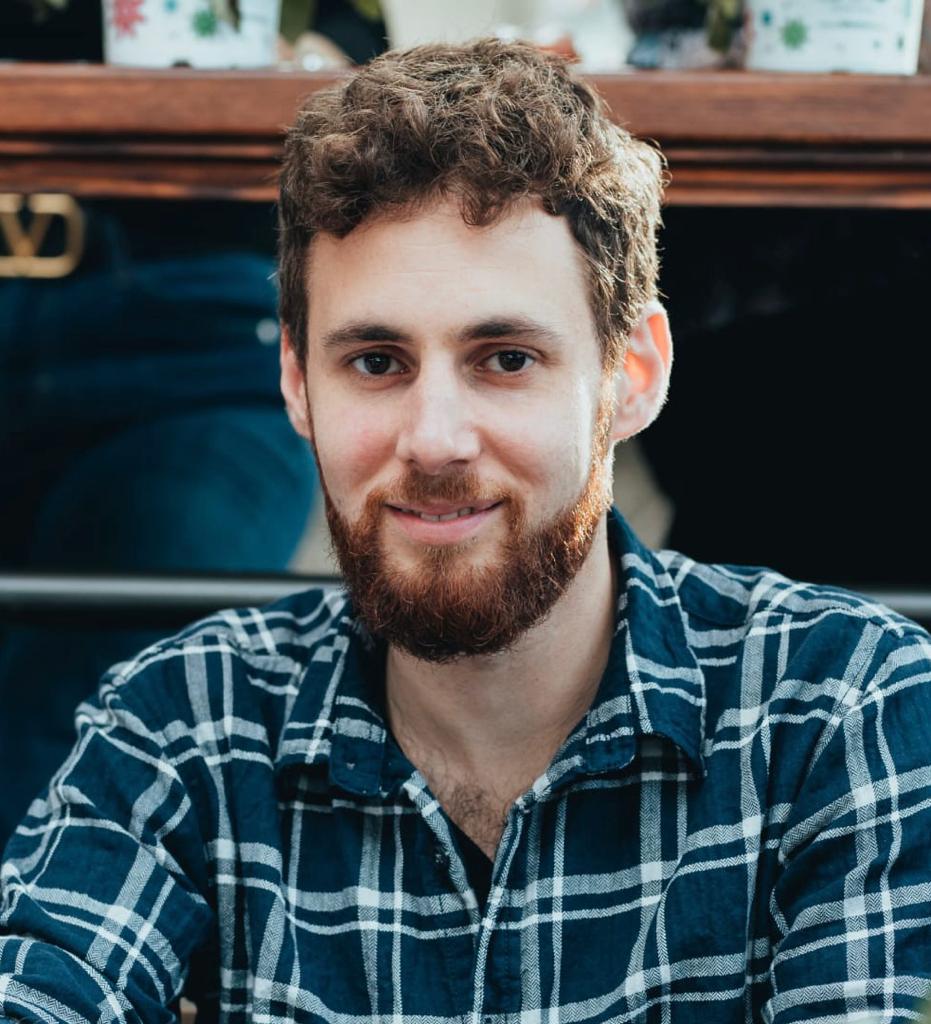}}]{Hanan Libhaber} is an M.Sc. student in the SISE Department at BGU. His research interests include machine learning and cyber-security. Contact him at hananlib@post.bgu.ac.il.
\end{IEEEbiography}

\begin{IEEEbiography}[{\includegraphics[width=1in,height=1.25in,clip,keepaspectratio]{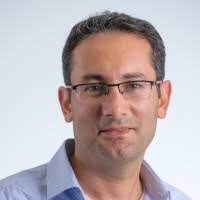}}]{Asaf Shabtai} is a professor in the SISE Department at BGU. His research interests include computer and network security, and machine learning. He received a Ph.D. in information systems from BGU. Contact him at shabtaia@bgu.ac.il.
\end{IEEEbiography}

\appendix

\subsection{Features in our publicly available dataset}\label{appendix:features}

The following is a brief description of the features in our publicly available dataset:
\begin{enumerate}[leftmargin=*]

\item \textbf{IPFIX identifiers}
\begin{itemize}
\item device\_id: An identifier of the inspected device from which this flow emanated (0-6 represent the devices deployed in the home networks, 7 represents the device in our lab)
\item flow\_start\_day: The day of the first packet of this flow
\item flow\_start\_hour: The hour of the first packet of this flow
\item flow\_start\_minute: The minute of the first packet of this flow
\item flow\_start\_second: The second of the first packet of this flow
\item flow\_start\_millisecond: The millisecond of the first packet of this flow
\item source\_network\_id: An identifier of the home network from which this flow emanated (0-4 represent the home networks, 5 represents the network in our lab)
\end{itemize}

\item \textbf{Raw IPFIX features}~\cite{ipfixspec}
\begin{itemize}
\item protocol\_identifier: The protocol number in the IP packet header, which identifies the IP packet payload type (6 represents TCP, 17 represents UDP)
\item flow\_duration\_milliseconds: The difference in time [milliseconds] between the first observed packet of this flow and the last observed packet of this flow
\item octet\_delta\_count: The number of octets sent in this flow, including IP header(s) and IP payload
\item packet\_delta\_count: The number of packets sent in this flow
\item avg\_packet\_size: The average size of packets in this flow, calculated as $octet\_delta\_count / packet\_delta\_count$
\item flow\_end\_reason: The reason for this flow's termination, e.g., idle timeout, TCP FIN flag, etc.
\item tcp\_control\_bits: The TCP control bits observed for the packets of this flow (SYN, ACK, FIN, PSH, RST, URG), encoded as a bit field (1 if present in any of this flow's packets, 0 otherwise)
\end{itemize}

\item \textbf{Features calculated using the raw features of the same IPFIX}
\begin{itemize}
\item network\_class\_of\_destination\_IP\_address: Determines which part of the IP address belongs to the network prefix and which part belongs to the host suffix (see~\cite{babatunde2014comparative})
\item network\_prefix\_of\_destination\_IP\_address\_anonimized: An anonimized prefix of the destination IP address, which indicates~\cite{liu2020tpii,meidan2020deNAT} the network of the host
\item inter\_arrival\_time\_milliseconds: The time [milliseconds] between the previous flow's start time and this flow's start time, calculated for this device\_id as $flow\_start_{(t)}-flow\_start_{(t-1)}$
\end{itemize}

\item \textbf{Reputation intelligence features}~\cite{webroot}
\begin{itemize}
\item reputation\_status: A flag indicating whether the destination IP address is suspected of being involved in any of the following categories of malicious activity: spam, windows exploits, web attacks, botnets, scanners, DoS, reputation, phishing, proxy, mobile threats, or tor proxy.
\end{itemize}

\item \textbf{`Time-based' features} (calculated using the full complete hour preceding flow\_start)
\begin{itemize}
\item same\_dest\_port\_count\_pool: A collaborative feature indicating the number of times in which the destination port was contacted by any of the monitored devices (identified by their device\_id) within the said time window
\item same\_dest\_IP\_count\_pool: A collaborative feature indicating the number of times in which network\_prefix\_of\_destination\_IP\_address\_anonimized was contacted by any of the monitored devices within the given time window 
\end{itemize}

\item \textbf{DNS-related features}
\begin{itemize}
\item has\_DNS\_request\_from\_pool: A (collaborative) flag indicating whether a preceding DNS request that can be associated with this flow was captured 
\item DNS\_host\_percentage\_of\_numerical\_chars\_from\_pool: Percentage of numerical characters in the requested DNS host
\end{itemize}

\item \textbf{Experiment-specific features}
\begin{itemize}
\item actual\_label: The actual label of this flow (either `assumed benign' or any of the attacks we implemented in our lab)
\item partition: The subset to which this flow belonged in our experiments (either training, validation, or test)
\end{itemize}

\end{enumerate}

\subsection{Nomenclature}\label{appendix:abbreviations}

The following is a list of the acronyms we used in this paper (presented in alphabetical order), as well as their definitions.

\nomenclature{AE}{Autoencoder} 
\nomenclature{ARIMA}{Autoregressive Integrated Moving Average}
\nomenclature{AUPRC}{Area under the precision-recall curve}
\nomenclature{BIRCH}{Balanced Iterative Reducing and Clustering using Hierarchies} 
\nomenclature{C\&C}{Command and control}
\nomenclature{CADeSH}{(The name of our proposed method) Collaborative Anomaly Detection for IoT attack detection in Smart Homes} 
\nomenclature{CIDS}{Collaborative intrusion detection system} 
\nomenclature{DBSCAN}{Density-Based Spatial Clustering of Applications with Noise} 
\nomenclature{DDoS}{Distributed denial-of-service} 
\nomenclature{DL}{Deep learning} 
\nomenclature{DNN}{Deep Neural Network} 
\nomenclature{DPI}{Deep packet inspection} 
\nomenclature{FL}{Federated learning} 
\nomenclature{FP}{False positive} 
\nomenclature{FPR}{False positive rate} \nomenclature{GPU}{Graphics processing unit} 
\nomenclature{GRU}{Gated Recurrent Unit} 
\nomenclature{IAT}{Interarrival time} 
\nomenclature{IDS}{Intrusion detection system} 
\nomenclature{IF}{Isolation Forest} 
\nomenclature{IoT}{Internet of things} 
\nomenclature{IPFIX}{Internet Protocol Flow Information Export} 
\nomenclature{ISP}{Internet service provider} 
\nomenclature{LOF}{Local Outlying Factor} 
\nomenclature{LSTM}{Long Short-Term Memory} 
\nomenclature{MCOD}{Multi-Level Outlier Detection} 
\nomenclature{ML}{Machine learning} 
\nomenclature{MLP}{Multilayer Perceptron} 
\nomenclature{MSE}{Mean squared error}
\nomenclature{NIDS}{Network intrusion detection system} 
\nomenclature{Nmap}{Network Mapper}
\nomenclature{OCSVM}{One-Class Support Vector Machines} 
\nomenclature{OPTICS}{Ordering Points to Identify the Clustering Structure} 
\nomenclature{PCA}{Principal Component Analysis}  
\nomenclature{SDN}{Software defined network}  
\nomenclature{VSL}{Virtual state layer} 
\printnomenclature[1.4cm] 

\end{document}